\documentclass[sigconf]{acmart}

\AtBeginDocument{%
  }

\copyrightyear{2024}
\acmYear{2024}
\setcopyright{rightsretained}
\acmConference[MM '24]{Proceedings of the 32nd ACM International Conference on Multimedia}{October 28-November 1, 2024}{Melbourne, VIC, Australia}
\acmBooktitle{Proceedings of the 32nd ACM International Conference on Multimedia (MM '24), October 28-November 1, 2024, Melbourne, VIC, Australia}
\acmDOI{xxxxxxxxxxxxx}
\acmISBN{xxxxxxxxxxxxxxxxxxxx}

\makeatletter
\gdef\@copyrightpermission{
   \begin{minipage}{0.3\columnwidth}
     \href{https://creativecommons.org/licenses/by-nc-sa/4.0/}{\includegraphics[width=0.90\textwidth]{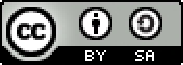}}
   \end{minipage}\hfill
   \begin{minipage}{0.7\columnwidth}
     \href{https://creativecommons.org/licenses/by-nc-sa/4.0/}{This work is licensed under a Creative Commons Attribution-ShareAlike International 4.0 License.}
   \end{minipage}
   \vspace{5pt}
}
\makeatother

\usepackage{multirow}
\usepackage{graphicx}
\usepackage{color}
\usepackage{algorithm}
\usepackage{algpseudocode}
\usepackage{subcaption}
\usepackage{pifont}

\begin{document}

\title{Data Generation Scheme for Thermal Modality with Edge-Guided Adversarial Conditional Diffusion Model}

\author{Guoqing Zhu}
\affiliation{
  \institution{Harbin Institute of Technology, Shenzhen}
  \city{Shenzhen}
  \country{China}}
\email{guoqingzhu1996@gmail.com}

\author{Honghu Pan}
\affiliation{
  \institution{University of Macau}
  \city{Macau}
  \country{China}}
\email{honghupan@um.edu.mo}

\author{Qiang Wang}
\authornote{Corresponding author}
\affiliation{
\institution{Harbin Institute of Technology, Shenzhen}
\city{Shenzhen}
\country{China}}
\email{qiang.wang@hit.edu.cn}

\author{Chao Tian}
\affiliation{
\institution{Harbin Institute of Technology, Shenzhen}
\city{Shenzhen}
\country{China}}
\email{tianchao@stu.hit.edu.cn}

\author{Chao Yang}
\affiliation{
\institution{Harbin Institute of Technology, Shenzhen}
\city{Shenzhen}
\country{China}}
\email{20b951014@stu.hit.edu.cn}

\author{Zhenyu He}
\authornotemark[1]
\affiliation{
\institution{Harbin Institute of Technology, Shenzhen}
\city{Shenzhen}
\country{China}}
\email{zhenyuhe@hit.edu.cn}

\renewcommand{\shortauthors}{Guoqing Zhu et al.}

\begin{abstract}
In challenging low-light and adverse weather conditions, thermal vision algorithms, especially object detection, have exhibited remarkable potential, contrasting with the frequent struggles encountered by visible vision algorithms. Nevertheless, the efficacy of thermal vision algorithms driven by deep learning models remains constrained by the paucity of available training data samples.
To this end, this paper introduces a novel approach termed the edge-guided conditional diffusion model (ECDM). This framework aims to produce meticulously aligned pseudo thermal images at the pixel level, leveraging edge information extracted from visible images. 
By utilizing edges as contextual cues from the visible domain, the diffusion model achieves meticulous control over the delineation of objects within the generated images. To alleviate the impacts of those visible-specific edge information that should not appear in the thermal domain, a two-stage modality adversarial training (TMAT) strategy is proposed to filter them out from the generated images by differentiating the visible and thermal modality. 
Extensive experiments on LLVIP demonstrate ECDM’s superiority over
existing state-of-the-art approaches in terms of image generation quality.
The pseudo thermal images generated by ECDM also help to boost the performance of various thermal object detectors by up to 7.1 mAP\@. Code is available at \href{https://github.com/lengmo1996/ECDM}{\textcolor{red}{https://github.com/lengmo1996/ECDM}}.
\end{abstract}

\ccsdesc[500]{Computing methodologies~Neural networks}
\ccsdesc[300]{Computing methodologies~Object detection}

\keywords{Diffusion model, Thermal image generation, Thermal object detection}

\maketitle

\section{Introduction}
In scenarios characterized by low-light or dark conditions, visible sensors often fail to yield substantial information, whereas thermal sensors capitalize on thermal radiation and temperature differentials. This sensitivity renders them particularly proficient in detecting temperature-related entities, notably livings and vehicles, within obscured settings. The superiority of thermal vision motivates numerous studies~\cite{CNNBased,SaliencyMap,BottomUp,UCIG,MLS} dedicated to thermal vision applications, particularly thermal object detection, and yield noteworthy enhancements in this domain.
\begin{figure}[t]
  \centering
  \begin{subfigure}[b]{0.8in}
      \centering
      \includegraphics[width=0.8in]{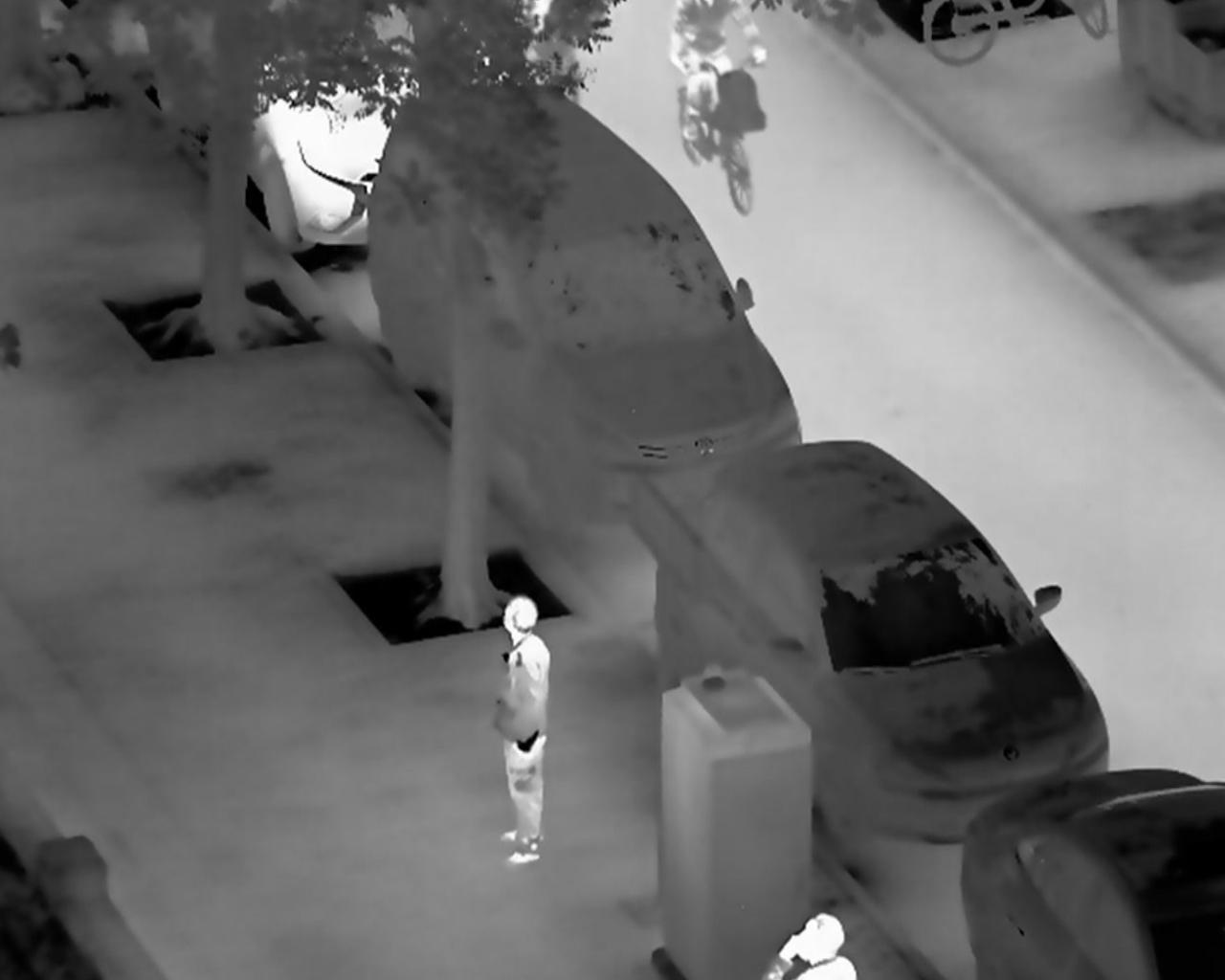}
      \caption{Ground truth}\label{Fig:gt}
  \end{subfigure}
  \begin{subfigure}[b]{0.8in}
      \centering
      \includegraphics[width=0.8in]{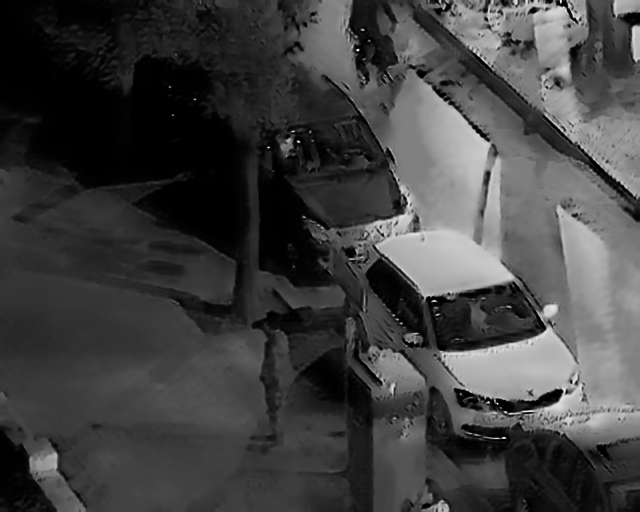}
      \caption{UGATIT}\label{Fig:ugatit}
  \end{subfigure}
  \begin{subfigure}[b]{0.8in}
      \centering
      \includegraphics[width=0.8in]{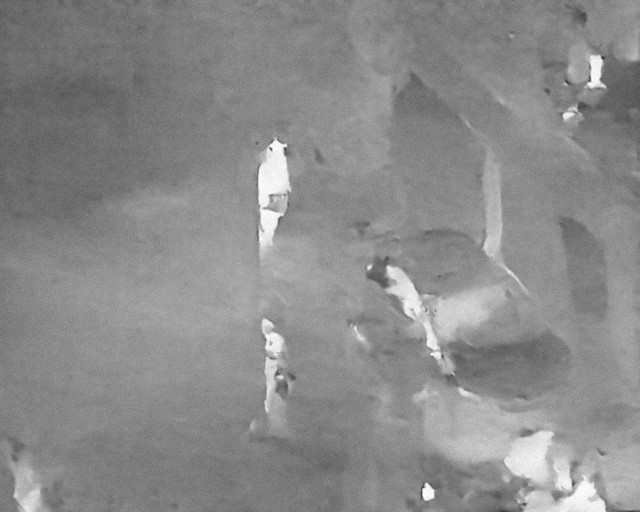}
      \caption{DDIM}\label{Fig:ddim}
  \end{subfigure}
  \begin{subfigure}[b]{0.8in}
      \centering
      \includegraphics[width=0.8in]{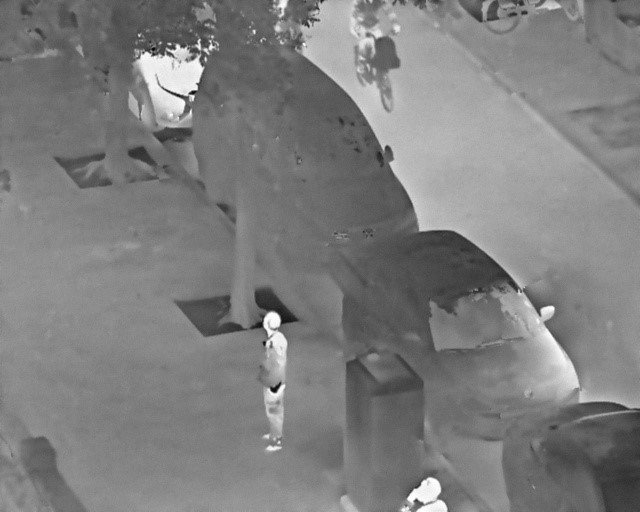}
      \caption{ECDM (Ours)}\label{Fig:ecdm}
  \end{subfigure}
  \caption{A sample comparison of generated thermal images between different methods and ground truth.\ (\subref{Fig:gt}) A ground truth thermal image, (\subref{Fig:ugatit}) a generated thermal image by UGATIT, (\subref{Fig:ddim}) a generated thermal image by DDIM and (\subref{Fig:ecdm}) a generated thermal image by ECDM (Ours). }\label{Fig:vistir}
  \Description[A sample comparison of generated thermal images between different methods and ground truth.]{A sample comparison of generated thermal images between different methods and ground truth.\ (\subref{Fig:gt}) A ground truth thermal image, (\subref{Fig:ugatit}) a generated thermal image by UGATIT, (\subref{Fig:ddim}) a generated thermal image by DDIM and (\subref{Fig:ecdm}) a generated thermal image by ECDM (Ours). }
\end{figure}

However, the efficacy of thermal vision applications remains notably curtailed by the paucity of available training samples. 
For example, the LLVIP dataset~\cite{LLVIP} only contains a mere 12,000 training thermal images, constituting only one-ninth of the visible samples contained within the COCO dataset~\cite{COCO}.
The procurement of expansive training data and the meticulous labeling of precise annotations necessitate extensive human labor and substantial time investment. To address this challenge, extant studies mainly harness two methodologies for augmenting training datasets to facilitate deep model training: 3D synthesis and deep generative models.
The 3D synthesis methods~\cite{FakeThermalImage,ThermalSynth} commence by generating a subset of 3D objects, followed by the application of a rudimentary thermal shader to render these objects, thereby engendering synthetic thermal images. Nevertheless, the images produced by the latest thermal sensor simulators still exhibit significant disparities when compared to those captured using real equipments.
Recent studies in the domain of generative models, particularly within the realm of Generative Adversarial Networks (GANs)~\cite{pix2pixGAN,UI2IT}, serves as an impetus for the generation of training data for thermal object detection~\cite{ImageGenarationForThermal,RobustSynth}.
However, existing GAN-based methods necessitate the availability of paired visible and thermal images for training deep models. The constraint often proves challenging to meet in practical contexts.
\begin{figure}[t]
\centering
\includegraphics[width=2.8in]{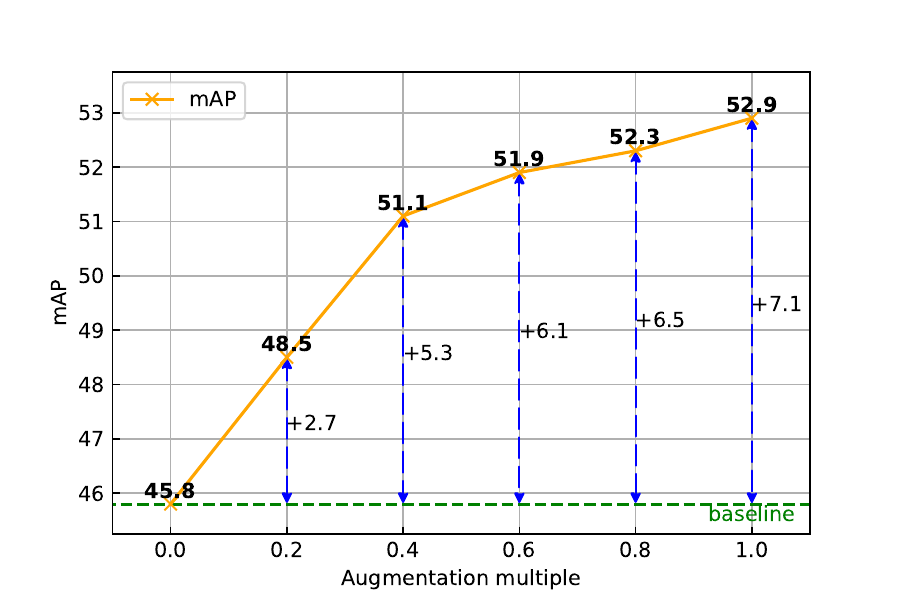}
\caption{The performance of RetinaNet trained with various amounts of generated pseudo training data. The x-axis indicates the augmentation multiple. For example, 0.2 indicates that the generated pseudo training data in the entire training sample is only 20\% of the real data.}\label{Fig:argmulti}
\Description[The performance of RetinaNet trained with various amounts of generated pseudo training data.]{The performance of RetinaNet trained with various amounts of generated pseudo training data. The x-axis indicates the augmentation multiple. For example, 0.2 indicates that the generated pseudo training data in the entire training sample is only 20\% of the real data.}
\end{figure}

For mitigating the scarcity of thermal data, this study delves into the applicability of the diffusion model to the task of generating thermally-aligned pixel-level images.
To this end, we propose an edge-guided adversarial condition diffusion model (ECDM) for thermal data generation. The basic idea of ECDM is learning the conditional probabilistic density of thermal images under the condition of the given visible image edge information. By incorporating the idea of adversarial learning to ease detrimental impacts stemming from extraneous and irrelevant edge details in the visible domain. 
As illustrated in Figure~\ref{Fig:ecdm}, our ECDM can simultaneously reconstruct object shape and object thermal radiation characteristics while other methods are only good in one aspect. As shown in Figure~\ref{Fig:ugatit}, GAN-based methods often tend to reconstruct object shapes but introduce irrelevant edges and abnormal details in thermal modality. As shown in Figure~\ref{Fig:ddim}, other Diffusion-based methods often tend to reconstruct object thermal radiation characteristics.

In summary, the main contributions of ECDM are threefold:
\begin{enumerate}
  \item ECDM engenders pixel-level thermally-aligned images by the generation process, bypassing the necessity for annotated visible-thermal pairs. This innovation augments the available datasets for thermal object detection by effectively generating thermal images from visible images.
  \item We devise a conditional diffusion model to estimate the conditional probabilistic density of thermal images under the constraints of the given visible image edges. Furthermore, we develop an adversarial training strategy to filter out the extraneous edge information from the visible domain that should not appear in the thermal domain.
  \item Extensive experiments on LLVIP demonstrate ECDM’s superiority over
existing state-of-the-art approaches in terms of image generation quality. The applicability of ECDM in generating training samples is also evaluated on the classical object detection task, wherein ECDM brings up to 7.1\% mAP improvement for the detectors  (as shown in Figure~\ref{Fig:argmulti}). 
\end{enumerate}

\section{Related work}\label{Sec:RelatedWork}
To solve the lacking of thermal images, some studies attempt to employ the domain adaptation techniques~\cite{BottomUp}, by fine-tuning the pretrained visible object detectors into the thermal domain. 
The main promising ideas are multi-level feature alignments~\cite{UnsupervisedT2I} and style consistency constraints~\cite{StyleConsistency}.
Nonetheless, while domain adaptation methods may mitigate the issue of insufficient annotations, they continue to face challenges in the absence of thermal images.

Another mainstream of studies rely on generative-based methods to generate synthetic thermal images. In~\cite{SynthHMI,FakeThermalImage,ThermalSynth}, there is a discussion of the use of virtual environments to create synthetic thermal images. These methods rely on intricate 3D models now focusing only on objects rather than whole scenes and employ infrared physics-based rendering. 
In~\cite{BorrowFromAnywhere,RobustSynth}, generative models are discussed to create synthetic thermal images.
But these generative models-based methods can not generate pixel-level alignments of thermal images from visible images. 

Deep generative models (DGMs) are neural networks trained to approximate the probability distributions of data. After training successfully, we can generate new samples from the underlying distribution. Generative Adversarial Networks (GANs)~\cite{GAN}, as a type of DGMs, have been extensively employed in the image-to-image translation tasks~\cite{CycleGAN,pix2pixGAN,LPTN,CVIIT}. Basic GANs consist of a generator and a discriminator under an adversarial training framework. The adversarial training process can be modeled as a min-max game. However, they have drawbacks such as poor convergence characteristics, especially on the thermal modality with rare textures.\

Recently, diffusion models (DMs)~\cite{DDPM} as a novel paradigm in the generative model, were shown impressive generative capabilities in high level of details~\cite{ReviewDiffusionModelVision}. Compared to GANs, this approach has a more stable training process and produces a greater range of diverse images. 
Recent advancements in DMs have demonstrated the ability to control the generation process, including details, through various conditions like image~\cite{Palette,BBDM,EGSDE}, class~\cite{DMBeatGAN}, and text~\cite{StableDiffusion}. DMs and their variants possess intriguing properties, such as stable training, generative diversity in images, and details control through conditions. These properties make them suitable for generating data from visible images for the thermal object detection task. 

Diffusion-GAN~\cite{DiffusionGAN} attempt utilize the advantage of the flexible diffusion model to stability the training process of GANs. In contrast to~\cite{DiffusionGAN} which injects adaptive noise via diffusion at various time steps to provide higher training stability over strong GAN baselines, our two-stage modality adversarial training (TMAT) strategy utilizes adversarial training to mitigate the distribution mismatch in generating images under diverse conditions.

\begin{figure*}[t]
  \centering
  \includegraphics[width=7in]{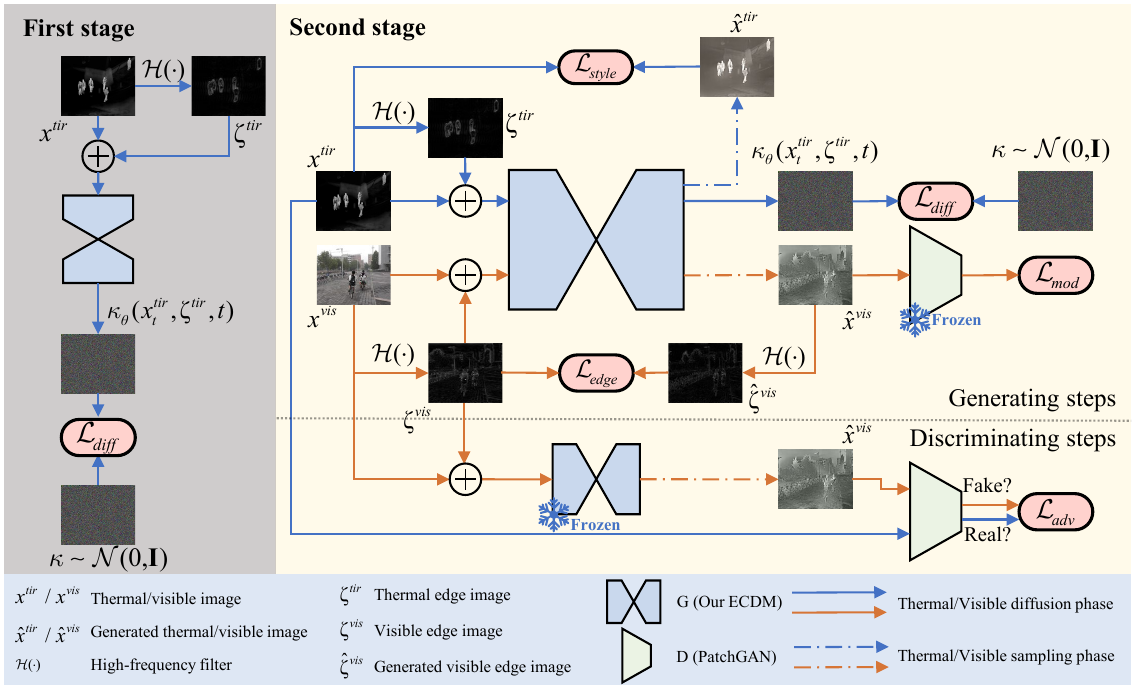}
  \caption{Illustration of our Two-stage Modality Adversarial Training (TMAT) strategy. During the first stage, we only use $x^{tir}$ as input and train the ECDM to learn the distribution of $p_{model}(x^{tir}|\zeta ^{tir})$. In the second stage, we use unpaired $x^{vis}$ and $x^{tir}$ as input and utilize GANs to reduce the gap between visible and thermal domains. This helps us learn the distribution of $p_{model}(x^{tir}|\zeta ^{vis})$ for approximating $p_{tir}(x^{tir})$.}\label{Fig:TMAT}
  \Description[Illustration of our Two-stage Modality Adversarial Training (TMAT) strategy.]{Illustration of our Two-stage Modality Adversarial Training (TMAT) strategy. During the first stage, we only use $x^{tir}$ as input and train the ECDM to learn the distribution of $p_{model}(x^{tir}|\zeta ^{tir})$. In the second stage, we use unpaired $x^{vis}$ and $x^{tir}$ as input and utilize GANs to reduce the gap between visible and thermal domains. This helps us learn the distribution of $p_{model}(x^{tir}|\zeta ^{vis})$ for approximating $p_{tir}(x^{tir})$.}
\end{figure*}
\section{Methodology}\label{Sec:Methodology}
\subsection{Framework Overview}\label{Subsec:Framework}
In this section, we first formalize the problem of \textit{generating pseudo training data for thermal object detection}. 
As visible object detection datasets typically exhibit greater scale than their thermal counterparts, we leverage existing visible datasets to craft pseudo training samples for thermal object detection.
Given a real visible object detection dataset $\mathcal{D}^{vis} =\{ x^{vis}_i,y^{vis}_i \} ^N_{i=1}$ contains $N$ visible images and a dataset $\mathcal{D}^{tir} =\{ x^{tir}_i,y^{tir}_i \} ^M_{i=1}$ contains $M$ real thermal infrared images, where each $x^{vis}_i$, $x^{tir}_i$ is an image sampled from a distribution $p_{vis}(x^{vis})$ or $p_{tir}(x^{tir})$, respectively. The corresponding annotations for each image are labeled as $y^{vis}_i$ and $y^{tir}_i$. In training generative models, the goal is to learn a model distribution $p_{model}(x^{tir}|x^{vis})$ that matches $p_{tir}(x^{tir})$.

However, different from the task of image generation or image-to-image translation, our generated pseudo images are prepared for thermal object detection. Consequently, generative thermal images necessitate pixel-level alignment with their corresponding visible images to accurately represent objects. To attain this precise alignment, we introduce the Edge Condition Diffusion Model (ECDM) rooted in conditional diffusion models. In our approach, edge images play a crucial role as the guiding condition during the sampling process for the creation of training samples.

Although edge information can bridge the thermal and visible domains, some discrepancies persist between the corresponding edge images in these domains. To address this challenge, we introduce a two-stage modality adversarial training strategy instead of a direct end-to-end training approach for the ECDM. Initially, we utilize thermal edge images to train the ECDM, enabling it to translate thermal edge images into thermal images. Subsequently, we leverage the trained ECDM as a generator and devise a discriminator. Through adversarial training resembling a GAN approach, we work towards minimizing the disparity between synthetically generated thermal images under visible edge conditions and authentic thermal images.
\begin{figure}[t]
  \centering
  \includegraphics[width=2.8in]{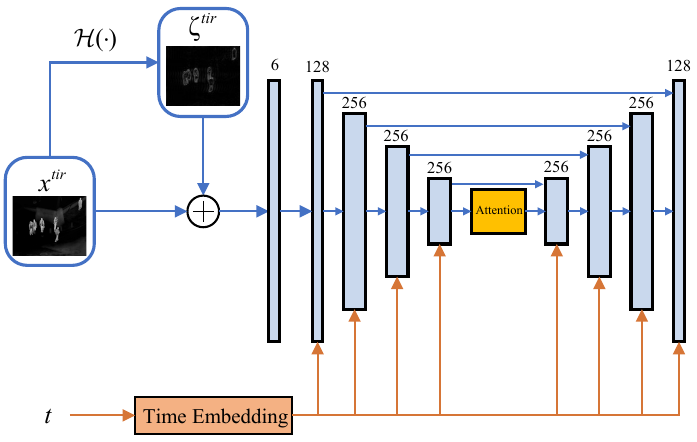}
  \caption{The architecture of ECDM\@. The numbers over the light blue rectangle blocks denote the channels of feature maps. The yellow rectangle block denotes an attention layer.}\label{Fig:ECDM}
  \Description[The architecture of ECDM\@.]{The architecture of ECDM\@. The numbers over the light blue rectangle blocks denote the channels of feature maps. The yellow rectangle block denotes an attention layer.}
\end{figure}

\subsection{Edge-guided Conditional Diffusion Model}\label{Subsec:ECDM}
Texture, shape, and color stand as the paramount visual cues in visual recognition~\cite{ShapeTextureColor}. However, due to the substantial differences in primary radiation, a considerable gap exists between thermal infrared and visible images. Specifically, thermal images have lack color information and exhibit lower texture than their visible counterparts. Shape information exhibits notable similarity between thermal and visible images within the same scene. We posit that this shape information serves as a bridge, mitigating the gap between the thermal and visible domains to some degree. The shape information in the images can be roughly extracted by high-frequency filtering
\begin{equation}
  \zeta ^m_i=\mathcal{H} (x^m_i),
\end{equation}
where $m\in \{ vis, tir \}$ is a modality indicator superscript, and $\mathcal{H} (\cdot)$ is an edge extracted operator contains a fast Fourier transform, a high pass filtering, and an inverse fast Fourier transform. The extracted shape information is also termed as edge images. Different from image generation task in vanilla diffusion models~\cite{DDPM,DDIM,DiffusionGAN} or conditioned on the class label, text, or natural images, the proposed ECDM introduces prior knowledge related to the fine-granularity content of objects, i.e., edge images. 

The ECDM comprises both the diffusion process and the reverse process. It destroys the input thermal image $x^{tir}_0 \in  \mathcal{D}^{tir}$ to a standard Gaussian noise $x^{tir}_T\sim\mathcal{N} (0,\mathbf{I} )$ by gradually adding small Gaussian noise in $T$ diffusion steps in forward process. The forward process of ECDM is presented as follows
\begin{equation}
  q(x^{tir}_{1:T}|x^{tir}_0)=\prod_{t=1}^{T}q(x^{tir}_{t}|x^{tir}_{t-1}),
\end{equation}
\begin{equation}
  q(x^{tir}_{t}|x^{tir}_{t-1})\sim\mathcal{N} (x^{tir}_{t};\sqrt{1-\beta_t}x^{tir}_{t-1},\beta_t \mathbf{I}),
\end{equation}
where $\beta_t$ is a small positive constant to control the variants of the added noise in diffusion step $t( 0\leq t\leq T)$. Following\cite{DDPM}, the variance schedule is predefined linearly increasing from $\beta_1=10^{-4}$ to $\beta_T=0.02$ and diffusion steps is seted $T=1000$. The noised sample at diffusion step $t$ can be directly calculated by
\begin{equation}
  x^{tir}_{t}=\sqrt{\bar{\alpha}_t}x^{tir}_{0} +\sqrt{1-\bar{\alpha}_t}\kappa, \kappa \sim\mathcal{N} (0,\mathbf{I} ),
\end{equation}
where $\alpha_t=1-\beta_t, \bar{\alpha}_t=\prod^t_{i=1}\alpha_i$.

The reverse process of ECDM is conditioned on the edge images to bridge the gap between the thermal domain and the visible domain while capturing the fine-granularity content of objects. Note that the different training stages use edge images in different domains. The reverse process is written as 
\begin{equation}
  p_\theta(x^{tir}_{0:T-1}|x^{tir}_{T},\zeta ^m)=\prod_{t=1}^{T}p_\theta(x^{tir}_{t-1}|x^{tir}_{t},\zeta ^m),
\end{equation}
\begin{equation}
  p_\theta(x^{tir}_{t-1}|x^{tir}_{t},\zeta ^m)\sim \mathcal{N} (x^{tir}_{t-1};\mu_\theta(x^{tir}_{t},\zeta ^m,t),\sigma ^2_t\mathbf{I}).
\end{equation}
The parameterizations of $\mu_\theta$ and $\sigma_\theta$ are defined by 
\begin{equation}
  \label{eq:muE}
  \mu_\theta(x^{tir}_{t},\zeta ^m,t)=\frac{1}{\sqrt{\bar{\alpha}_t}}\Bigr(x^{tir}_{t}-\frac{\beta_t}{\sqrt{1-\bar{\alpha}_t}}\kappa_\theta(x^{tir}_{t},\zeta ^m,t)\Bigr),
\end{equation}
\begin{equation}
  \sigma^2_t=\frac{1-\bar{\alpha}_{t-1}}{1-\bar{\alpha}_{t}}\beta_t,
\end{equation}
where $\kappa_\theta$ is a neural network parameterized by $\theta$, implemented by a modified UNet~\cite{UNet,DDPM}. In our work, $\kappa_\theta$ takes the noised image $x^{tir}_{t}$, the conditional edge image $\zeta ^m$, and the diffusion step $t$ as input. The the diffusion step $t$ are fed into a Transformer sinusoidal position embedding layer~\cite{Transformer} with a given embedding dimension, followed by a Linear + sigmoid layer. Then, for each downsample and upsample block, an additional Linear + sigmoid layer is employed to align the channel dimensions of the feature maps. Figure~\ref{Fig:ECDM} shows the architecture of ECDM\@.

\subsection{Two-stage Modality Adversarial Training}\label{Subsec:TMAT}
Our goal is to learn the distribution of $p_{model}(x^{tir}|\zeta ^{vis})$. However, attempting to manually train the ECDM in a straightforward end-to-end manner, such as $\zeta ^m=\zeta ^{vis}$ in~\eqref{eq:muE}, while theoretically feasible, becomes challenging in practice due to the substantial divergence between the thermal and visible domains. 

To address this challenge, we propose a two-stage modality adversarial training strategy. As illustrated in Figure~\ref{Fig:TMAT}, in the first stage, we set $\zeta ^m=\zeta ^{tir}$ and train the model to learn the distribution of $p_{model}(x^{tir}|\zeta ^{tir})$. The condition $\zeta ^{tir}$ and generative images $x^{tir}$ all remain within the thermal domain, simplifying the distribution learning process. In the second stage, we incorporate the principles of GANs, employing the ECDM previously trained in the first stage as the generator and employing a thermal modality authenticity indicator as the discriminator. At this stage, we begin by generating images from the generator under a distinct condition $\zeta ^m=\zeta ^{vis}$, which introduces modality bias due to the incongruity between $p(\zeta ^{tir})$ and $p(\zeta ^{vis})$. We then mitigate this modality bias through adversarial training. The complete two-stage modality adversarial training procedure is detailed in Algorithm\ref{Algo:TMAT}.
\begin{algorithm}[t]
  \caption{Two-stage Modality Adversarial Training}\label{Algo:TMAT}
  \begin{algorithmic}[1]
    \Require{}the thermal image $x^{tir}$, the visible images $x^{vis}$, the first stage training epochs $S_{diff}$, the second stage training epochs $S_{adv}$, the steps of generating $S_G$, the steps of discriminating $S_D$.
    \State{}$x_{T}\sim \mathcal{N} (0,\mathbf{I} )$
    \State{}$\zeta ^{tir}=\mathcal{H} (x^{tir})$, $\zeta ^{vis}=\mathcal{H} (x^{vis})$
    \For{$i=0$ to $S_{diff}$} 
    \State{}Update $\kappa _\theta$ by descending its gradient:
    \State{}$\triangledown _{\kappa _\theta}\parallel \kappa -\kappa _\theta(x^{tir}_{t},\zeta ^{tir},t) \parallel ^2_2$
    \EndFor{}

    \State{}$\hat{x}^{tir}\leftarrow G (\kappa,\zeta ^{tir})$, $\hat{x}^{vis}\leftarrow G (\kappa,\zeta ^{vis})$

    \For{$j=0$ to $S_{adv}$} 
    \For{$k=0$ to $S_G$}
    \State{}Update $\kappa _\theta$ by descending its gradient:
    \State{}$\triangledown _{\kappa _\theta}\parallel \kappa -\kappa _\theta(x^{tir}_{t},\zeta ^{tir},t) \parallel ^2_2$
    \State{}$\triangledown _{\kappa _\theta}\Vert \hat{x}^{tir}-x^{tir}\Vert ^2_2$
    \State{}$\triangledown _{\kappa _\theta}\Vert 1- D(\hat{x}^{vis})\Vert ^2_2$
    \State{}$\triangledown _{\kappa _\theta}\Vert \mathcal{H} (x^{vis})-\mathcal{H} (\hat{x}^{vis})\Vert ^2_2$
    \EndFor{}
    \For{$l=0$ to $S_D$}
    \State{}Update $D$ by descending its gradient:
    \State{}$\triangledown _{D}[\lambda_{real}\Vert\log(D (x ^{tir}))\Vert ^2_2$
    \State{}\qquad$+$\;$\Vert\log(1-D (\hat{x}^{vis}))\Vert ^2_2]$
    \EndFor{}
    \EndFor{}
    \State{}\textbf{\Return{}}ECDM ($\kappa _\theta$).
  \end{algorithmic}
\end{algorithm}

Specifically, in the first stage, we focus on training the ECDM by optimizing the standard variational bound on negative log-likelihood. Following the reparameterization trick in~\cite{DDPM}, the training objective of ECDM in the first stage training process is
\begin{equation}
  \mathcal{L} _{diff}=\mathbb{E} _{x^{tir}_0,\kappa,t}\parallel \kappa -\kappa _\theta(x^{tir}_{t},\zeta ^{tir},t) \parallel ^2_2.
\end{equation}

In the second stage, we leverage the ECDM as a generator $G $ and introduce a discriminator $D $ by PatchGAN~\cite{pix2pixGAN}, where $G $ aims to generate synthetic thermal images $\hat{x}^{vis}\sim p_{tir}(x^{tir})$ the condition of visible edge images, $D$ aims to distinguish the real and synthetic thermal images. To ensure that the generated thermal images are indistinguishable from authentic ones, we employ an adversarial loss~\cite{GAN}
\begin{equation}
  \begin{split}
    \mathcal{L} _{adv}=&\lambda_{real}{\mathbb{E} _{x ^{tir}}}[\log(D (x ^{tir}))]+ \\
    &\mathbb{E} _{\kappa,\zeta ^{vis}}[\log(1-D (G (\kappa,\zeta ^{vis})  ))],
  \end{split}
\end{equation}
where $\lambda_{real}$ is a hyper-parameter to balance the adversarial loss components. We set $\lambda_{real}=10$ in our experiments.
To further narrow the gap between the generated thermal images and actual ones, we incorporate a style-consistency loss defined as
\begin{equation}
  \mathcal{L} _{style}=\Vert \hat{x}^{tir}-x^{tir}\Vert ^2_2,
\end{equation}
and design a modality-consistency loss
\begin{equation}
  \mathcal{L} _{mod}=\Vert 1- D(\hat{x}^{vis})\Vert ^2_2,
\end{equation}
where $\hat{x}^{tir}:=G (\kappa,\zeta ^{tir})$ and $\hat{x}^{vis}:=G (\kappa,\zeta ^{vis})$ represent synthetic thermal images under conditions of thermal edge images or visible edge images, respectively. Additionally, we utilize an edge loss to preserve accurate boundaries and highly detailed shapes, which is defined as
\begin{equation}
  \mathcal{L} _{edge}=\Vert \mathcal{H} (x^{vis})-\mathcal{H} (\hat{x}^{vis})\Vert ^2_2.
\end{equation}

Finally, we express the objective functions to optimize $G$ and $D$, respectively, as follows
\begin{equation}
  \begin{split}
    \mathcal{L}_G=&\lambda_{diff}\mathcal{L} _{diff}+\lambda_{style}\mathcal{L} _{style}+\\
    &\lambda_{mod}\mathcal{L} _{mod}+\lambda_{edge}\mathcal{L} _{edge},
  \end{split}
\end{equation}
\begin{equation}
    \mathcal{L}_D=\mathcal{L}_{adv},
\end{equation}
where $\lambda_{diff}, \lambda_{style}, \lambda_{mod},\lambda_{edge}$ are hyper-parameters to control the weight of different loss. We use $\lambda_{diff}=0.1,\lambda_{style}=100,\lambda_{mod}=1,\lambda_{edge}=1000$ in all our experiments. Moreover, we use dpm-solver++~\cite{DPMSolverPP} to improve the sampling speed of ECDM\@. The sampling parameters are listed in the supplementary material.

\section{Experiments}\label{Sec:Experiments}
\subsection{Experimental Settings}
\subsubsection{Datasets}
Our experiments are mainly on the LLVIP dataset \cite{LLVIP}, FLIR dataset~\cite{FLIR}, and Person Re-identification in the Wild Dataset (PRW)\cite{PRW}.  
LLVIP consists of 15,488 pairs of visible-thermal images, captured under low-light conditions using a binocular surveillance camera. These paired images are precisely aligned both spatially and temporally. For brevity, we refer to this dataset as$\mathcal{D} _{llvip}$.
FLIR has two versions: v1.3 (2019) and v2.0 (2021). We utilize v2.0 for our experiments. Our analysis focuses on five categories: person, bike, car, light, and sign.
PRW comprises 11,816 frames captured during the summer months using a visible camera. These frames were extracted from the Market-1501 dataset~\cite{Market151}. This dataset is annotated for both person re-identification and pedestrian detection tasks. We refer to this dataset as $\mathcal{D} _{prw}$.

\subsubsection{Metrics}
We use FID~\cite{FIDGAN}, LPIPS~\cite{LPIPS}, PSNR, SSIM, and KID\cite{KID} to measure the quality of generated thermal images. We denote the implementation in~\cite{PytorhcFID} as FID\@. Besides, the FID implemented in~\cite{CleanFID} is FID-C, and the FID implemented using CLIP instead of InceptionV3 is FID-C$_{clip}$. Additionally, we employ KID, a metric similar to FID but with a polynomial kernel for an unbiased estimator.

We use standard mean Average Percision~\cite{COCO} (mAP) under different Intersection over Union (IoU) thresholds as the metrics to measure the gain of generated pseudo training data to the performance of thermal detectors.

\subsection{Implementation Details}\label{Subsec:Implementation}

We train the ECDM on four NVIDIA 3090 24GB GPUs, utilizing a batch size of 4 and resizing the input images to a resolution of $512 \times 640$. The generator $G$ and discriminator $D$ are optimized using Adam with $\beta_1=0.9$ and $\beta_2=0.999$. The learning rate is set to $0.00002$. Our training procedure involves setting $S_{diff}=70$, $S_{adv}=20$, $S_G=2$, and $S_D=1$. Further details can be found in the supplementary material.

Our training process in Sec.~\ref{subsec:showcase} utilizes two NVIDIA 3090 24GB GPUs with a batch size of 32 and employs SGD as the optimizer. We set the base learning rate to 0.0002, momentum to 0.99, and weight decay to 0.0001. To ensure a fair comparison, we train them for 40k iterations and use a multistep learning rate scheduler with steps at 12000, 18000, and 32000 iterations.

\subsection{Quantitative and Qualitative Comparison to Visible-to-thermal Translation Task}
We demonstrate that ECDM can deliver competitive results in visible-to-thermal translation tasks. We compare our method with several state-of-the-art methods: pix2pixGAN~\cite{pix2pixGAN}, CycleGAN~\cite{CycleGAN}, UGATIT~\cite{UGATIT}, LPTN~\cite{LPTN}, 
VSAIT~\cite{VSAIT}, DDIM~\cite{DDIM}, BBDM~\cite{BBDM} and DDBM~\cite{DDBM}. We also benchmark some energy-based/flow-based models in our experiments, but their performance was inferior to the GAN-based methods presented in Table 1. We report the performance under $512\times640$ resolution which specifically addresses thermal object detections. For DDIM, we modified the original architecture as shown in Figure~\ref{Fig:ECDM}. However, in our case, we used visible images from the LLVIP dataset as the conditioning input for the visible-to-thermal image translation task. We evaluate the performance of BBDM and DDBM by upsampling generated images from 256$\times$256, due to its difficulty in converging at 512$\times$640.

As can be seen in Table~\ref{Tab:SOTA}, ECDM achieves superior performance to SOTA thermal image generators on the LLVIP dataset. It suggests that by introducing visible edge conditions to guide diffusion modeling, our proposed ECDM effectively generates high-quality pixel-level aligned pseudo thermal images with visible edge images.
\begin{table}[htb]
  \centering
  \caption{Quantitative comparison on the LLVIP dataset. 
  $^\S$ indicates upsampling from the 256$\times$256 resolution. Best results highlighted in \textbf{bold}, second best in \underline{underline}.}\label{Tab:SOTA}
  \resizebox{3.35in}{!}{
    \begin{tabular}{l c ccc}
    \toprule
    Method &  FID$\downarrow$ &LPIPS$\downarrow$ & SSIM$\uparrow$ & PSNR$\uparrow$ \\
    \midrule
    pix2pixGAN (CVPR 2017)& 317.38 &0.474 &0.211&11.251 \\
    CycleGAN (ICCV 2017)&183.80 &\underline{0.354}& 0.283& 12.196\\
    UGATIT (ICLR 2020)&  \underline{178.71}&0.359 &0.285 &12.970 \\
    LPTN (CVPR 2021)&209.84  &0.396 &0.245 &11.658 \\
    VSAIT (ECCV 2022)&211.30 &0.360 &0.277 &\underline{13.050} \\
    \midrule
    DDIM (ICLR 2021)&  325.87 &0.454&\underline{0.393}&11.741\\
    BBDM$^\S$ (CVPR 2023)& 265.06 &0.436 &0.311 &11.728 \\
    DDBM$^\S$ (ICLR 2024)&423.21&0.494&0.251&12.235\\ 
    \midrule
    ECDM (Ours)&\textbf{139.91} &\textbf{0.141}&\textbf{0.507}&\textbf{13.130}\\
    \bottomrule
  \end{tabular}
  }
\end{table}

\begin{figure*}[t]
  \centering
  \includegraphics[width=7in]{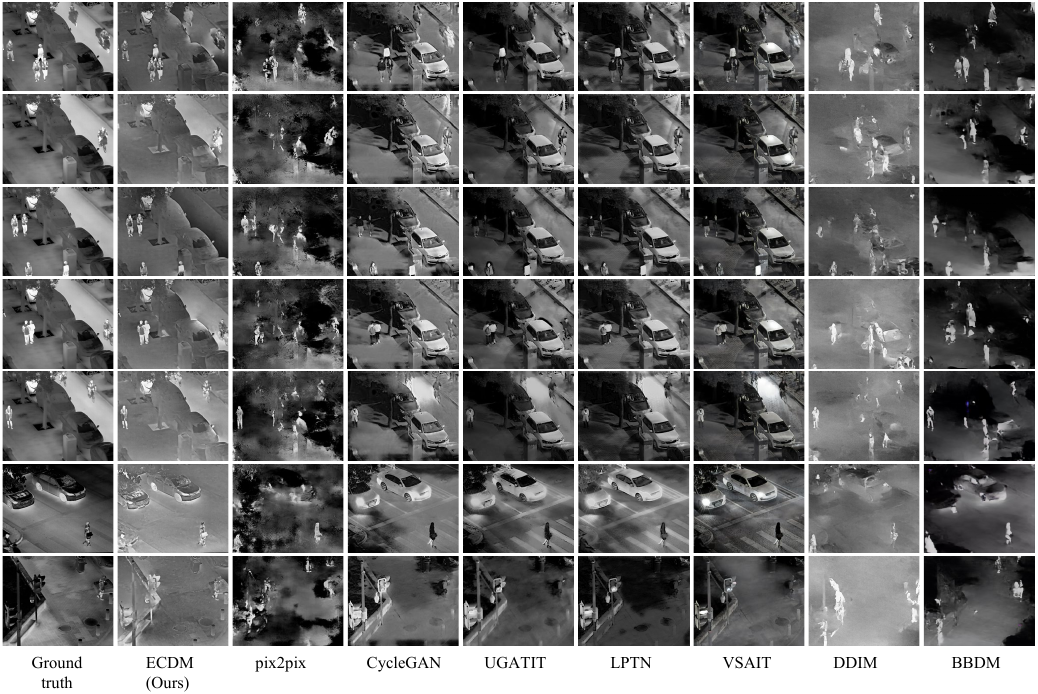}
  \caption{Qualitative comparison of our proposed method with other state-of-the-art methods on the LLVIP test dataset. To ensure fairness and randomness, we use Python's random module with a fixed seed (1234) to select four images from the dataset. The selected images are `190145.jpg', `190345.jpg', `190373.jpg', `190405.jpg', `190480.jpg', `220224.jpg', `260261.jpg'. More visual results can be found in supplementary material.}\label{Fig:com}
  \Description[Qualitative comparison of our proposed method with other state-of-the-art methods on the LLVIP test dataset.]{Qualitative comparison of our proposed method with other state-of-the-art methods on the LLVIP test dataset. To ensure fairness and randomness, we use Python's random module with a fixed seed (1234) to select four images from the dataset. The selected images are `190145.jpg', `190345.jpg', `190373.jpg', `190405.jpg', `190480.jpg', `220224.jpg', `260261.jpg'. More visual results can be found in supplementary material.}
\end{figure*}

We also show qualitative comparison with other methods in Figure~\ref{Fig:com}. Our qualitative comparison is based on two key principles: \textbf{object shape reconstruction} and \textbf{object thermal radiation characteristics reconstruction}.

In Figure 5, although some methods can reconstruct full objects like cars, they still fail to capture the thermal radiation characteristics of objects. For example, the tires of moving vehicles and the exposed skin of pedestrians should appear brighter in thermal images. GAN-based methods, such as CycleGAN and UGATIT, excel in object shape reconstruction but fall short in object thermal radiation characteristics reconstruction. These GAN-based methods also tend to reconstruct the object texture rather than the object thermal radiation characteristics. Conversely, diffusion-based methods, such as DDIM and BBIM, demonstrate proficiency in thermal radiation characteristics reconstruction but struggle with shape reconstruction. Our proposed method stands out by simultaneously achieving reconstruction of both shape and thermal radiation characteristics. More visual results can be found in supplementary material. 

\subsection{Model Complexity Comparison}
We compare the complexity of ECDM with other methods in the number of parameters (\#params) and FLOPs. In Table~\ref{Tab:complexity}, our method is comparable to GAN-based method except LPTN in \#params. Compared with DDIM, our method introduces an additional discriminator, resulting in an increase of 2.765M \#params.  Diffusion-based methods generally require higher FLOPs because of multisteps reversal process compared with GAN-based methods. However, our method benefits from the DPM-Solver++, which results in the lowest FLOPs among diffusion-based methods.

\begin{table}[htb]
	\centering
	\caption{Comparison of model complexity and parameters. \#params denotes the number of parameters. G and T after the values represent the unit of FLOPs. Different colors are used for better distinguish.}\label{Tab:complexity}
	  \begin{tabular}{cl cc}
		\toprule
	  Type & Method & \#params (M) $\downarrow$ & FLOPs $\downarrow$\\
	  \midrule
    \multirow{5}*{GAN-based}& pix2pixGAN &60.290 &44.598\textcolor{blue}{G} \\
	  &CycleGAN&28.286 &496.415\textcolor{blue}{G} \\
	  &UGATIT&32.946 &134.577\textcolor{blue}{G} \\
    &LPTN &0.871 &13.629\textcolor{blue}{G} \\
	  &VSAIT&65.492 &642.878\textcolor{blue}{G} \\
    \midrule
    \multirow{3}*{Diffusion-based}&DDIM &34.431 & 733.246\textcolor{red}{T}\\
	  &BBDM&273.095 &806.380\textcolor{red}{T} \\
    &ECDM (Ours) &37.196 &120.986\textcolor{red}{T} \\
	  \bottomrule
	\end{tabular}
\end{table}
We also report that generating $512\times640$ of one thermal image needs about 14.7s on a NVIDIA 3090 GPU\@. However, we can apply some acceleration schemes of diffusion to tackle the efficiency problem. Besides, the data generation is a one-time operation, and any other downstream tasks can use them.

\subsection{Transferability of ECDM}
In the experimental setup described above, the visible edge images and target generated thermal images are from the same dataset (all in the LLVIP dataset), resulting in a small gap between them. In practice, generating pseudo thermal images that can yield gains for downstream tasks such as thermal object detection poses a challenging problem: will the generated training data be useful when the conditions are far from the target domain? This raises the issue of model transferability.

\begin{table}[htb]
  \centering
  \caption{Ablation study for transferability of ECDM.}\label{Tab:condition2}
    \begin{tabular}{l l cccc}
    \toprule
    \multicolumn{2}{c}{Condition} & \multirow{2}{*}{FID$\downarrow$} & \multirow{2}{*}{FID-C$\downarrow$} & \multirow{2}{*}{FID-C$_{clip}\downarrow$} & \multirow{2}{*}{KID$\downarrow$} \\
    \cmidrule{0-1}
    Training & Sampling& & & & \\
    \midrule
    $\mathcal{D} _{llvip}^{tir}$&$\mathcal{D} _{prw}$& 306.66&305.00&66.59&0.2881\\
    $\zeta_{llvip}^{tir}$&$\zeta_{prw}$& 249.49&245.84&36.84&0.2265\\
    $\mathcal{D} _{llvip}^{vis}$&$\mathcal{D} _{prw}$& 267.59&264.56&41.43&0.2744\\ 
    $\zeta_{llvip}^{vis}$&$\zeta_{prw}$&   278.38&280.99&44.75&0.3198\\ 
    $\mathcal{D} _{prw}$&$\mathcal{D} _{prw}$& 294.32&285.56&38.85&0.3255\\
    $\zeta_{prw}$&$\zeta_{prw}$& 266.97&267.29&44.35&0.2997\\
    \bottomrule
  \end{tabular}

\end{table}

To evaluate the transferability of ECDM, we train our model under various conditions, including thermal image conditions (denoted as $\mathcal{D}_{llvip}^{tir}$), thermal edge conditions (denoted as $\zeta_{llvip}^{tir}$), visible images in the LLVIP dataset (denoted as $\mathcal{D} _{llvip}^{vis}$), visible edge images in the LLVIP dataset (denoted as $\zeta_{llvip}^{vis}$), and visible images in the PRW dataset (denoted as $\mathcal{D} _{prw}$), visible edge images in the PRW dataset (denoted as $\zeta_{prw}$). We directly sample thermal images under $\mathcal{D} _{prw}$ or $\zeta_{prw}$ conditions from the aforementioned cases. As shown in Table~\ref{Tab:condition2}, substantial degradations in various metrics are evident due to domain and dataset disparities. Nonetheless, well-trained ECDM attains a minimal FID-C score under cross-domain and cross-dataset sampling conditions.

\subsection{Effects of TMAT with Different Edge Information}

To validate the effectiveness of our TMAT strategy in the face of transferability, we train the ECDM with TMAT under diverse edge inputs. 
As shown in Table~\ref{Tab:TMAT}, the trends of various metrics are consistent, demonstrating that TMAT diminishes the gap between disparate domains and datasets.

\begin{table}[htb]
  \centering
  \caption{Ablation study for TMAT.}\label{Tab:TMAT}
  \resizebox{3.35in}{!}{
    \begin{tabular}{l c cccc}
    \toprule
    Condition & TMAT&FID$\downarrow$ & FID-C$\downarrow$ & FID-C$_{clip}\downarrow$ & KID$\downarrow$ \\
    \midrule
    \multirow{2}{*}{Thermal edge images}&\ding{56}&249.49&245.84&36.84&0.2265\\
    &\ding{52}&248.36&241.96&38.98&0.2292\\
    \midrule
    \multirow{2}{*}{Visible edge images (night time)}&\ding{56}&278.38&280.99&44.75&0.3198\\
    &\ding{52}&264.70&259.63&37.02&0.2499\\
    \midrule
    \multirow{2}{*}{Visible edge images (day time)}&\ding{56}&266.97&267.29&44.35&0.2997\\
    &\ding{52}&258.59&251.82&43.18&0.2485\\
    \bottomrule
  \end{tabular}
} 
\end{table}

\subsection{A Showcase of ECDM on Thermal Object Detection}\label{subsec:showcase}
To investigate the impact of the number of pseudo thermal images on thermal object detection performance, we select RetinaNet~\cite{RetinaNet} as our baseline and train it under the same settings, except for the training data. The training data consists of two parts: real thermal images from the training set of LLVIP and pseudo thermal images generated by our ECDM using PRW edge images. The number of real thermal images is fixed in the training data and the number of pseudo thermal images generated by our ECDM is controled by the augmentation multiple ratio. For instance, an augmentation multiple ratio of 0.2 signifies that the generated pseudo training data constitutes only 20\% of the real data in the entire training set. We experiment with diverse augmentation multiple ratios, namely 0, 0.2, 0.4, 0.6, 0.8, and 1.0, and observe the impact on mAP\@. As depicted in Figure~\ref{Fig:argmulti}, the mAP improves gradually from 0 to 7.1, with the most significant enhancement occurring at a ratio of 1.0.

To further effective the generalization of pseudo training data generated by our ECDM, we train various object detectors, including Faster RCNN~\cite{FasterRCNN}, RetinaNet~\cite{RetinaNet}, CenterNet~\cite{CenterNet}, VFNet~\cite{VFNet}, and DINO~\cite{DINO}. For a fair comparison, we maintain an augmentation multiple ratio of 1.0 throughout this experiment. 

Our generated training data yield mAP improvements ranging from 1.1 (CenterNet) to 7.1 (RetinaNet) across different detectors on the LLVIP dataset. Notably, most detectors, excluding VFNet, exhibit mAP improvements between 0.7 and 1.8, demonstrating the effectiveness of our pseudo training data on the FLIR dataset.

\begin{table}[htb]
  \centering
  \caption{Using pseudo data training different detectors on LLVIP dataset and FLIR dataset. The red color means performance improvement while the green color represents performance decrease.}\label{Tab:detectors}
  \resizebox{3.35in}{!}{
    \begin{tabular}{l cccc ccc}
    \toprule
    Method & Batch size&Backbone&Dataset&Pseudo data & mAP & mAP@50 & mAP@75 \\
    \midrule
    \multirow{4}{*}{Faster RCNN}&\multirow{4}{*}{32}&\multirow{4}{*}{Resnet-50}&\multirow{2}{*}{LLVIP}&\ding{56}&49.0&89.2&48.4\\
    &&&&\ding{52}&50.3 \textcolor{red}{(+1.3)}&89.2&51.6\\
    \cmidrule{4-8}
    &&&\multirow{2}{*}{FLIR}&\ding{56}&23.5&42.7&22.6\\
    &&&&\ding{52}&24.4 \textcolor{red}{(+0.9)}&44.0&23.1\\
    \midrule

    \multirow{4}{*}{RetinaNet}&\multirow{4}{*}{32}&\multirow{4}{*}{Resnet-50}&\multirow{2}{*}{LLVIP}&\ding{56}&45.8&90.3&40.7\\
    &&&&\ding{52}&52.5 \textcolor{red}{(+7.1)}&92.8&53.6\\
    \cmidrule{4-8}
    &&&\multirow{2}{*}{FLIR}&\ding{56}&14.5&29.4&12.4\\
    &&&&\ding{52}&15.2 \textcolor{red}{(+0.7)}&30.8&13.2\\
    \midrule

    \multirow{4}{*}{CenterNet}&\multirow{4}{*}{32}&\multirow{4}{*}{Resnet-50}&\multirow{2}{*}{LLVIP}&\ding{56}&53.4&91.3&56.3\\
    &&&&\ding{52}&54.5 \textcolor{red}{(+1.1)}&93.1&57.7\\
    \cmidrule{4-8}
    &&&\multirow{2}{*}{FLIR}&\ding{56}&25.5&48.9&22.8\\
    &&&&\ding{52}&27.3 \textcolor{red}{(+1.8)}&52.1&24.5\\
    \midrule
    
    \multirow{4}{*}{VFNet}&\multirow{4}{*}{32}&\multirow{4}{*}{Resnet-50}&\multirow{2}{*}{LLVIP}&\ding{56}&52.2&91.3&54.1\\
    &&&&\ding{52}&54.7 \textcolor{red}{(+2.5)}&92.9&58.6\\
    \cmidrule{4-8}
    &&&\multirow{2}{*}{FLIR}&\ding{56}&15.1&32.0&12.2\\
    &&&&\ding{52}&13.0 \textcolor{green}{(-2.1)}&27.8&10.7\\
    \midrule

    \multirow{4}{*}{DINO}&\multirow{4}{*}{2}&\multirow{4}{*}{Swin-L}&\multirow{2}{*}{LLVIP}&\ding{56}&40.2&72.9&39.8\\
    &&&&\ding{52}&44.2 \textcolor{red}{(+4.0)}&74.9&46.9\\
    \cmidrule{4-8}
    &&&\multirow{2}{*}{FLIR}&\ding{56}&7.6&16.0&6.5\\
    &&&&\ding{52}&9.2 \textcolor{red}{(+1.6)}&20.8&6.8\\

    \bottomrule
  \end{tabular}
}
\end{table}

The Varifocal Loss in VFNet focuses the training on those high-quality positive examples that are more important for achieving a higher AP than those low-quality ones~\cite{VFNet}. 
Considering the dataset characteristics, FLIR comprises objects of different scales, with a predominant presence of small-scale objects. LLVIP primarily consists of medium-scale objects, while PRW contains a mix of medium- and large-scale objects. The unique design of the Varifocal Loss in VFNet places more emphasis on the generated objects compared to other detectors since large-scale objects are more easily identified as positive examples than small-scale objects. This design also explains the performance decrease of VFNet on the FLIR dataset and its performance improvement on the LLVIP dataset.

We also evaluate the performance of pseudo-training data generated by ECDM or other image-to-image translation techniques on the RetinaNet. As shown in Table~\ref{Tab:SOTA2}, when compared to the absence of any additional training data, the utilization of generated pseudo-training data proves effective in enhancing the performance of RetinaNet. Notably, generated thermal images from other methods also contain useful semantic information for perception tasks, because they are practically reconstruct shape or thermal radiation characteristics. However, our method excels in achieving both shape and thermal radiation characteristics reconstruction simultaneously, leading to the highest improvement (+7.1 mAP) compared with other methods.

\begin{table}[htb]
  \centering
  \caption{Comparing the impact of pseudo training data generated by different methods on the performance of RetinaNet. `None' indicates that no generated training data is used, while `PRW' denotes the utilization of the PRW dataset as additional training data. Best results highlighted in \textbf{bold}, second best in \underline{underline}.
  }\label{Tab:SOTA2}
  \resizebox{3.35in}{!}{
    \begin{tabular}{l l cc}
    \toprule
    Generating Method & mAP &mAP@50 & mAP@75 \\
    \midrule
    None (baseline)&45.8&90.3&40.7\\
    PRW  &48.0\textcolor{red}{(+2.2)}&91.4&44.8\\
    CycleGAN      &51.9\textcolor{red}{(+6.1)}&92.6&52.9\\
    UGATIT        &51.1\textcolor{red}{(+5.3)}&89.7&53.8\\
    LPTN          &52.4\textcolor{red}{(+6.6)}&92.0&54.4\\
    VSAIT         &\underline{52.6}\textcolor{red}{(+6.8)}&93.0&54.6\\
    DDIM          &50.6\textcolor{red}{(+4.8)}  &92.4 &49.9\\
    ECDM (Ours)   &\textbf{52.9}\textcolor{red}{(+7.1)}&92.7&55.3\\
    \bottomrule
  \end{tabular}
  }
\end{table}

\section{Limitations and Future Works}
Our ECDM currently requires high-quality edge images, which limits its applicability in certain scenarios.
Furthermore, we also noticed that the generated images often have global color levels error, especially on the ground, traffic lights, and woods, which may account for the large FID score in Table~\ref{Tab:SOTA}. Additionally, as illustrated in Fig~\ref{Fig:com}, our method and DDIM tend to generate brighter thermal images compared to the ground truth or the CycleGAN result. Notably, BBDM, also a diffusion-based method, does not exhibit this trend. Considering the comparison of model parameters in Table~\ref{Tab:complexity}. We attribute this issue to the U-Net's limitations in parameter count and learning capacity. 

Considering the relationship between the thermal object detection and other downstream tasks which also suffer the lacking of training samples like thermal object tracking~\cite{DeepTIOT}, thermal semantic segmentation~\cite{TISemantic}, and other related task~\cite{GaitRec,ReID,TCDesc}, our method is expected to bring performance improvements to them. However, tracking tasks typically require image sequences as input, necessitating the design of additional modules to ensure frame-inter consistency for generated images. It also exhibits potential applicability to image-to-image translation where edges can act as connectors between the source and target domain, particularly when the texture information in the target domain is sparse.

\section{Conclusion}\label{Sec:Conclusion}
In this paper, we introduce a novel data generation scheme for the thermal modality, called ECDM, which leverages a diffusion model to generate pixel-level aligned thermal images. Our approach utilizes edge images extracted from visible images as a condition to guide the diffusion model in learning the fine control of object boundaries in the generated image. To address the domain gap between thermal and visible images, we propose TMAT, a method that trains our ECDM to generate thermal images from visible edge images. Our extensive experiments demonstrate the promising performance of ECDM, and we conduct an exhaustive ablation study to analyze its effectiveness.

\begin{acks}
This work was supported by the National Natural Science Foundation of China (No. 62172126), and the Shenzhen Research Council (No. JCYJ20210324120202006).
\end{acks}

\bibliographystyle{ACM-Reference-Format}
\bibliography{sample-base}


\begin{thebibliography}{55}


\ifx \showCODEN    \undefined \def \showCODEN     #1{\unskip}     \fi
\ifx \showDOI      \undefined \def \showDOI       #1{#1}\fi
\ifx \showISBNx    \undefined \def \showISBNx     #1{\unskip}     \fi
\ifx \showISBNxiii \undefined \def \showISBNxiii  #1{\unskip}     \fi
\ifx \showISSN     \undefined \def \showISSN      #1{\unskip}     \fi
\ifx \showLCCN     \undefined \def \showLCCN      #1{\unskip}     \fi
\ifx \shownote     \undefined \def \shownote      #1{#1}          \fi
\ifx \showarticletitle \undefined \def \showarticletitle #1{#1}   \fi
\ifx \showURL      \undefined \def \showURL       {\relax}        \fi
\providecommand\bibfield[2]{#2}
\providecommand\bibinfo[2]{#2}
\providecommand\natexlab[1]{#1}
\providecommand\showeprint[2][]{arXiv:#2}

\bibitem[Bińkowski et~al\mbox{.}(2021)]%
        {KID}
\bibfield{author}{\bibinfo{person}{Mikołaj Bińkowski}, \bibinfo{person}{Danica~J. Sutherland}, \bibinfo{person}{Michael Arbel}, {and} \bibinfo{person}{Arthur Gretton}.} \bibinfo{year}{2021}\natexlab{}.
\newblock \bibinfo{title}{Demystifying MMD GANs}.
\newblock
\newblock
\showeprint[arxiv]{1801.01401}~[stat.ML]


\bibitem[Blythman et~al\mbox{.}(2020)]%
        {SynthHMI}
\bibfield{author}{\bibinfo{person}{Richard Blythman}, \bibinfo{person}{Amr Elrasad}, \bibinfo{person}{Eoin O'Connell}, \bibinfo{person}{Paul Kielty}, \bibinfo{person}{Michael O'Byrne}, \bibinfo{person}{Mohamed Moustafa}, \bibinfo{person}{Cian Ryan}, {and} \bibinfo{person}{Joe Lemley}.} \bibinfo{year}{2020}\natexlab{}.
\newblock \showarticletitle{Synthetic thermal image generation for human-machine interaction in vehicles}. In \bibinfo{booktitle}{\emph{2020 Twelfth International Conference on Quality of Multimedia Experience (QoMEX)}}. IEEE, \bibinfo{pages}{1--6}.
\newblock


\bibitem[Bongini et~al\mbox{.}(2021)]%
        {FakeThermalImage}
\bibfield{author}{\bibinfo{person}{Francesco Bongini}, \bibinfo{person}{Lorenzo Berlincioni}, \bibinfo{person}{Marco Bertini}, {and} \bibinfo{person}{Alberto Del~Bimbo}.} \bibinfo{year}{2021}\natexlab{}.
\newblock \showarticletitle{Partially fake it till you make it: mixing real and fake thermal images for improved object detection}. In \bibinfo{booktitle}{\emph{Proceedings of the 29th ACM International Conference on Multimedia}}. \bibinfo{pages}{5482--5490}.
\newblock


\bibitem[Croitoru et~al\mbox{.}(2023)]%
        {ReviewDiffusionModelVision}
\bibfield{author}{\bibinfo{person}{Florinel-Alin Croitoru}, \bibinfo{person}{Vlad Hondru}, \bibinfo{person}{Radu~Tudor Ionescu}, {and} \bibinfo{person}{Mubarak Shah}.} \bibinfo{year}{2023}\natexlab{}.
\newblock \showarticletitle{Diffusion models in vision: A survey}.
\newblock \bibinfo{journal}{\emph{IEEE Transactions on Pattern Analysis and Machine Intelligence}} (\bibinfo{year}{2023}).
\newblock


\bibitem[Devaguptapu et~al\mbox{.}(2019)]%
        {BorrowFromAnywhere}
\bibfield{author}{\bibinfo{person}{Chaitanya Devaguptapu}, \bibinfo{person}{Ninad Akolekar}, \bibinfo{person}{Manuj M~Sharma}, {and} \bibinfo{person}{Vineeth N~Balasubramanian}.} \bibinfo{year}{2019}\natexlab{}.
\newblock \showarticletitle{Borrow from anywhere: Pseudo multi-modal object detection in thermal imagery}. In \bibinfo{booktitle}{\emph{Proceedings of the IEEE/CVF Conference on Computer Vision and Pattern Recognition Workshops}}. \bibinfo{pages}{0--0}.
\newblock


\bibitem[Dhariwal and Nichol(2021)]%
        {DMBeatGAN}
\bibfield{author}{\bibinfo{person}{Prafulla Dhariwal} {and} \bibinfo{person}{Alexander Nichol}.} \bibinfo{year}{2021}\natexlab{}.
\newblock \showarticletitle{Diffusion models beat gans on image synthesis}.
\newblock \bibinfo{journal}{\emph{Advances in neural information processing systems}}  \bibinfo{volume}{34} (\bibinfo{year}{2021}), \bibinfo{pages}{8780--8794}.
\newblock


\bibitem[Duan et~al\mbox{.}(2019)]%
        {CenterNet}
\bibfield{author}{\bibinfo{person}{Kaiwen Duan}, \bibinfo{person}{Song Bai}, \bibinfo{person}{Lingxi Xie}, \bibinfo{person}{Honggang Qi}, \bibinfo{person}{Qingming Huang}, {and} \bibinfo{person}{Qi Tian}.} \bibinfo{year}{2019}\natexlab{}.
\newblock \showarticletitle{Centernet: Keypoint triplets for object detection}. In \bibinfo{booktitle}{\emph{Proceedings of the IEEE/CVF international conference on computer vision}}. \bibinfo{pages}{6569--6578}.
\newblock


\bibitem[FLIR(2019)]%
        {FLIR}
\bibfield{author}{\bibinfo{person}{Teledyne FLIR}.} \bibinfo{year}{2019}\natexlab{}.
\newblock \bibinfo{title}{Free Teledyne FLIR thermal dataset for algorithm training}.
\newblock \bibinfo{howpublished}{\url{https://www.flir.com/oem/adas/adas-dataset-form}}.
\newblock
\newblock
\shownote{Accessed:2023-08-01}.


\bibitem[Ge et~al\mbox{.}(2022)]%
        {ShapeTextureColor}
\bibfield{author}{\bibinfo{person}{Yunhao Ge}, \bibinfo{person}{Yao Xiao}, \bibinfo{person}{Zhi Xu}, \bibinfo{person}{Xingrui Wang}, {and} \bibinfo{person}{Laurent Itti}.} \bibinfo{year}{2022}\natexlab{}.
\newblock \showarticletitle{Contributions of shape, texture, and color in visual recognition}. In \bibinfo{booktitle}{\emph{European Conference on Computer Vision}}. Springer, \bibinfo{pages}{369--386}.
\newblock


\bibitem[Ghose et~al\mbox{.}(2019)]%
        {SaliencyMap}
\bibfield{author}{\bibinfo{person}{Debasmita Ghose}, \bibinfo{person}{Shasvat~M Desai}, \bibinfo{person}{Sneha Bhattacharya}, \bibinfo{person}{Deep Chakraborty}, \bibinfo{person}{Madalina Fiterau}, {and} \bibinfo{person}{Tauhidur Rahman}.} \bibinfo{year}{2019}\natexlab{}.
\newblock \showarticletitle{Pedestrian detection in thermal images using saliency maps}. In \bibinfo{booktitle}{\emph{Proceedings of the IEEE/CVF Conference on Computer Vision and Pattern Recognition Workshops}}. \bibinfo{pages}{0--0}.
\newblock


\bibitem[Goodfellow et~al\mbox{.}(2014)]%
        {GAN}
\bibfield{author}{\bibinfo{person}{Ian Goodfellow}, \bibinfo{person}{Jean Pouget-Abadie}, \bibinfo{person}{Mehdi Mirza}, \bibinfo{person}{Bing Xu}, \bibinfo{person}{David Warde-Farley}, \bibinfo{person}{Sherjil Ozair}, \bibinfo{person}{Aaron Courville}, {and} \bibinfo{person}{Yoshua Bengio}.} \bibinfo{year}{2014}\natexlab{}.
\newblock \showarticletitle{Generative adversarial nets}.
\newblock \bibinfo{journal}{\emph{Advances in neural information processing systems}}  \bibinfo{volume}{27} (\bibinfo{year}{2014}).
\newblock


\bibitem[Herrmann et~al\mbox{.}(2018)]%
        {CNNBased}
\bibfield{author}{\bibinfo{person}{Christian Herrmann}, \bibinfo{person}{Miriam Ruf}, {and} \bibinfo{person}{J{\"u}rgen Beyerer}.} \bibinfo{year}{2018}\natexlab{}.
\newblock \showarticletitle{CNN-based thermal infrared person detection by domain adaptation}. In \bibinfo{booktitle}{\emph{Autonomous Systems: Sensors, Vehicles, Security, and the Internet of Everything}}, Vol.~\bibinfo{volume}{10643}. SPIE, \bibinfo{pages}{38--43}.
\newblock


\bibitem[Heusel et~al\mbox{.}(2017)]%
        {FIDGAN}
\bibfield{author}{\bibinfo{person}{Martin Heusel}, \bibinfo{person}{Hubert Ramsauer}, \bibinfo{person}{Thomas Unterthiner}, \bibinfo{person}{Bernhard Nessler}, {and} \bibinfo{person}{Sepp Hochreiter}.} \bibinfo{year}{2017}\natexlab{}.
\newblock \showarticletitle{Gans trained by a two time-scale update rule converge to a local nash equilibrium}.
\newblock \bibinfo{journal}{\emph{Advances in neural information processing systems}}  \bibinfo{volume}{30} (\bibinfo{year}{2017}).
\newblock


\bibitem[Ho et~al\mbox{.}(2020)]%
        {DDPM}
\bibfield{author}{\bibinfo{person}{Jonathan Ho}, \bibinfo{person}{Ajay Jain}, {and} \bibinfo{person}{Pieter Abbeel}.} \bibinfo{year}{2020}\natexlab{}.
\newblock \showarticletitle{Denoising diffusion probabilistic models}.
\newblock \bibinfo{journal}{\emph{Advances in neural information processing systems}}  \bibinfo{volume}{33} (\bibinfo{year}{2020}), \bibinfo{pages}{6840--6851}.
\newblock


\bibitem[Isola et~al\mbox{.}(2017)]%
        {pix2pixGAN}
\bibfield{author}{\bibinfo{person}{Phillip Isola}, \bibinfo{person}{Jun-Yan Zhu}, \bibinfo{person}{Tinghui Zhou}, {and} \bibinfo{person}{Alexei~A Efros}.} \bibinfo{year}{2017}\natexlab{}.
\newblock \showarticletitle{Image-to-image translation with conditional adversarial networks}. In \bibinfo{booktitle}{\emph{Proceedings of the IEEE conference on computer vision and pattern recognition}}. \bibinfo{pages}{1125--1134}.
\newblock


\bibitem[Jia et~al\mbox{.}(2021)]%
        {LLVIP}
\bibfield{author}{\bibinfo{person}{Xinyu Jia}, \bibinfo{person}{Chuang Zhu}, \bibinfo{person}{Minzhen Li}, \bibinfo{person}{Wenqi Tang}, {and} \bibinfo{person}{Wenli Zhou}.} \bibinfo{year}{2021}\natexlab{}.
\newblock \showarticletitle{LLVIP: A visible-infrared paired dataset for low-light vision}. In \bibinfo{booktitle}{\emph{Proceedings of the IEEE/CVF International Conference on Computer Vision}}. \bibinfo{pages}{3496--3504}.
\newblock


\bibitem[Kieu et~al\mbox{.}(2021a)]%
        {BottomUp}
\bibfield{author}{\bibinfo{person}{My Kieu}, \bibinfo{person}{Andrew~D Bagdanov}, {and} \bibinfo{person}{Marco Bertini}.} \bibinfo{year}{2021}\natexlab{a}.
\newblock \showarticletitle{Bottom-up and layerwise domain adaptation for pedestrian detection in thermal images}.
\newblock \bibinfo{journal}{\emph{ACM Transactions on Multimedia Computing, Communications, and Applications (TOMM)}} \bibinfo{volume}{17}, \bibinfo{number}{1} (\bibinfo{year}{2021}), \bibinfo{pages}{1--19}.
\newblock


\bibitem[Kieu et~al\mbox{.}(2021b)]%
        {RobustSynth}
\bibfield{author}{\bibinfo{person}{My Kieu}, \bibinfo{person}{Lorenzo Berlincioni}, \bibinfo{person}{Leonardo Galteri}, \bibinfo{person}{Marco Bertini}, \bibinfo{person}{Andrew~D Bagdanov}, {and} \bibinfo{person}{Alberto Del~Bimbo}.} \bibinfo{year}{2021}\natexlab{b}.
\newblock \showarticletitle{Robust pedestrian detection in thermal imagery using synthesized images}. In \bibinfo{booktitle}{\emph{2020 25th International Conference on Pattern Recognition (ICPR)}}. IEEE, \bibinfo{pages}{8804--8811}.
\newblock


\bibitem[Kim et~al\mbox{.}(2020)]%
        {UGATIT}
\bibfield{author}{\bibinfo{person}{Junho Kim}, \bibinfo{person}{Minjae Kim}, \bibinfo{person}{Hyeonwoo Kang}, {and} \bibinfo{person}{Kwanghee Lee}.} \bibinfo{year}{2020}\natexlab{}.
\newblock \bibinfo{title}{U-GAT-IT: Unsupervised Generative Attentional Networks with Adaptive Layer-Instance Normalization for Image-to-Image Translation}.
\newblock
\newblock
\showeprint[arxiv]{1907.10830}~[cs.CV]


\bibitem[Li et~al\mbox{.}(2023)]%
        {BBDM}
\bibfield{author}{\bibinfo{person}{Bo Li}, \bibinfo{person}{Kaitao Xue}, \bibinfo{person}{Bin Liu}, {and} \bibinfo{person}{Yu-Kun Lai}.} \bibinfo{year}{2023}\natexlab{}.
\newblock \showarticletitle{BBDM: Image-to-image translation with Brownian bridge diffusion models}. In \bibinfo{booktitle}{\emph{Proceedings of the IEEE/CVF Conference on Computer Vision and Pattern Recognition}}. \bibinfo{pages}{1952--1961}.
\newblock


\bibitem[Liang et~al\mbox{.}(2021)]%
        {LPTN}
\bibfield{author}{\bibinfo{person}{Jie Liang}, \bibinfo{person}{Hui Zeng}, {and} \bibinfo{person}{Lei Zhang}.} \bibinfo{year}{2021}\natexlab{}.
\newblock \showarticletitle{High-resolution photorealistic image translation in real-time: A laplacian pyramid translation network}. In \bibinfo{booktitle}{\emph{Proceedings of the IEEE/CVF Conference on Computer Vision and Pattern Recognition}}. \bibinfo{pages}{9392--9400}.
\newblock


\bibitem[Lin et~al\mbox{.}(2017)]%
        {RetinaNet}
\bibfield{author}{\bibinfo{person}{Tsung-Yi Lin}, \bibinfo{person}{Priya Goyal}, \bibinfo{person}{Ross Girshick}, \bibinfo{person}{Kaiming He}, {and} \bibinfo{person}{Piotr Doll{\'a}r}.} \bibinfo{year}{2017}\natexlab{}.
\newblock \showarticletitle{Focal loss for dense object detection}. In \bibinfo{booktitle}{\emph{Proceedings of the IEEE international conference on computer vision}}. \bibinfo{pages}{2980--2988}.
\newblock


\bibitem[Lin et~al\mbox{.}(2014)]%
        {COCO}
\bibfield{author}{\bibinfo{person}{Tsung-Yi Lin}, \bibinfo{person}{Michael Maire}, \bibinfo{person}{Serge Belongie}, \bibinfo{person}{James Hays}, \bibinfo{person}{Pietro Perona}, \bibinfo{person}{Deva Ramanan}, \bibinfo{person}{Piotr Doll{\'a}r}, {and} \bibinfo{person}{C~Lawrence Zitnick}.} \bibinfo{year}{2014}\natexlab{}.
\newblock \showarticletitle{Microsoft coco: Common objects in context}. In \bibinfo{booktitle}{\emph{Computer Vision--ECCV 2014: 13th European Conference, Zurich, Switzerland, September 6-12, 2014, Proceedings, Part V 13}}. Springer, \bibinfo{pages}{740--755}.
\newblock


\bibitem[Liu and Ma(2022)]%
        {CVIIT}
\bibfield{author}{\bibinfo{person}{Hong Liu} {and} \bibinfo{person}{Lei Ma}.} \bibinfo{year}{2022}\natexlab{}.
\newblock \showarticletitle{Infrared Image Generation Algorithm Based on GAN and contrastive learning}. In \bibinfo{booktitle}{\emph{2022 International Conference on Artificial Intelligence and Computer Information Technology (AICIT)}}. IEEE, \bibinfo{pages}{1--4}.
\newblock


\bibitem[Liu et~al\mbox{.}(2017a)]%
        {UI2IT}
\bibfield{author}{\bibinfo{person}{Ming-Yu Liu}, \bibinfo{person}{Thomas Breuel}, {and} \bibinfo{person}{Jan Kautz}.} \bibinfo{year}{2017}\natexlab{a}.
\newblock \showarticletitle{Unsupervised image-to-image translation networks}.
\newblock \bibinfo{journal}{\emph{Advances in neural information processing systems}}  \bibinfo{volume}{30} (\bibinfo{year}{2017}).
\newblock


\bibitem[Liu et~al\mbox{.}(2021)]%
        {ImageGenarationForThermal}
\bibfield{author}{\bibinfo{person}{Peng Liu}, \bibinfo{person}{Fuyu Li}, \bibinfo{person}{Shanshan Yuan}, {and} \bibinfo{person}{Wanyi Li}.} \bibinfo{year}{2021}\natexlab{}.
\newblock \showarticletitle{Unsupervised image-generation enhanced adaptation for object detection in thermal images}.
\newblock \bibinfo{journal}{\emph{Mobile information systems}}  \bibinfo{volume}{2021} (\bibinfo{year}{2021}), \bibinfo{pages}{1--6}.
\newblock


\bibitem[Liu et~al\mbox{.}(2020)]%
        {MLS}
\bibfield{author}{\bibinfo{person}{Qiao Liu}, \bibinfo{person}{Xin Li}, \bibinfo{person}{Zhenyu He}, \bibinfo{person}{Nana Fan}, \bibinfo{person}{Di Yuan}, {and} \bibinfo{person}{Hongpeng Wang}.} \bibinfo{year}{2020}\natexlab{}.
\newblock \showarticletitle{Learning deep multi-level similarity for thermal infrared object tracking}.
\newblock \bibinfo{journal}{\emph{IEEE Transactions on Multimedia}}  \bibinfo{volume}{23} (\bibinfo{year}{2020}), \bibinfo{pages}{2114--2126}.
\newblock


\bibitem[Liu et~al\mbox{.}(2017b)]%
        {DeepTIOT}
\bibfield{author}{\bibinfo{person}{Qiao Liu}, \bibinfo{person}{Xiaohuan Lu}, \bibinfo{person}{Zhenyu He}, \bibinfo{person}{Chunkai Zhang}, {and} \bibinfo{person}{Wen-Sheng Chen}.} \bibinfo{year}{2017}\natexlab{b}.
\newblock \showarticletitle{Deep convolutional neural networks for thermal infrared object tracking}.
\newblock \bibinfo{journal}{\emph{Knowledge-Based Systems}}  \bibinfo{volume}{134} (\bibinfo{year}{2017}), \bibinfo{pages}{189--198}.
\newblock


\bibitem[Lu et~al\mbox{.}(2023)]%
        {DPMSolverPP}
\bibfield{author}{\bibinfo{person}{Cheng Lu}, \bibinfo{person}{Yuhao Zhou}, \bibinfo{person}{Fan Bao}, \bibinfo{person}{Jianfei Chen}, \bibinfo{person}{Chongxuan Li}, {and} \bibinfo{person}{Jun Zhu}.} \bibinfo{year}{2023}\natexlab{}.
\newblock \bibinfo{title}{DPM-Solver++: Fast Solver for Guided Sampling of Diffusion Probabilistic Models}.
\newblock
\newblock
\showeprint[arxiv]{2211.01095}~[cs.LG]


\bibitem[Madan et~al\mbox{.}(2023)]%
        {ThermalSynth}
\bibfield{author}{\bibinfo{person}{Neelu Madan}, \bibinfo{person}{Mia Sandra~Nicole Siemon}, \bibinfo{person}{Magnus~Kaufmann Gjerde}, \bibinfo{person}{Bastian~Starup Petersson}, \bibinfo{person}{Arijus Grotuzas}, \bibinfo{person}{Malthe~Aaholm Esbensen}, \bibinfo{person}{Ivan~Adriyanov Nikolov}, \bibinfo{person}{Mark~Philip Philipsen}, \bibinfo{person}{Kamal Nasrollahi}, {and} \bibinfo{person}{Thomas~B Moeslund}.} \bibinfo{year}{2023}\natexlab{}.
\newblock \showarticletitle{ThermalSynth: A Novel Approach for Generating Synthetic Thermal Human Scenarios}. In \bibinfo{booktitle}{\emph{Proceedings of the IEEE/CVF Winter Conference on Applications of Computer Vision}}. \bibinfo{pages}{130--139}.
\newblock


\bibitem[Marnissi et~al\mbox{.}(2022)]%
        {UnsupervisedT2I}
\bibfield{author}{\bibinfo{person}{Mohamed~Amine Marnissi}, \bibinfo{person}{Hajer Fradi}, \bibinfo{person}{Anis Sahbani}, {and} \bibinfo{person}{Najoua Essoukri~Ben Amara}.} \bibinfo{year}{2022}\natexlab{}.
\newblock \showarticletitle{Unsupervised thermal-to-visible domain adaptation method for pedestrian detection}.
\newblock \bibinfo{journal}{\emph{Pattern Recognition Letters}}  \bibinfo{volume}{153} (\bibinfo{year}{2022}), \bibinfo{pages}{222--231}.
\newblock


\bibitem[Munir et~al\mbox{.}(2022)]%
        {StyleConsistency}
\bibfield{author}{\bibinfo{person}{Farzeen Munir}, \bibinfo{person}{Shoaib Azam}, \bibinfo{person}{Muhammd~Aasim Rafique}, \bibinfo{person}{Ahmad~Muqeem Sheri}, \bibinfo{person}{Moongu Jeon}, {and} \bibinfo{person}{Witold Pedrycz}.} \bibinfo{year}{2022}\natexlab{}.
\newblock \showarticletitle{Exploring thermal images for object detection in underexposure regions for autonomous driving}.
\newblock \bibinfo{journal}{\emph{Applied Soft Computing}}  \bibinfo{volume}{121} (\bibinfo{year}{2022}), \bibinfo{pages}{108793}.
\newblock


\bibitem[Pan et~al\mbox{.}(2021)]%
        {TCDesc}
\bibfield{author}{\bibinfo{person}{Honghu Pan}, \bibinfo{person}{Yongyong Chen}, \bibinfo{person}{Zhenyu He}, \bibinfo{person}{Fanyang Meng}, {and} \bibinfo{person}{Nana Fan}.} \bibinfo{year}{2021}\natexlab{}.
\newblock \showarticletitle{TCDesc: Learning topology consistent descriptors for image matching}.
\newblock \bibinfo{journal}{\emph{IEEE Transactions on Circuits and Systems for Video Technology}} \bibinfo{volume}{32}, \bibinfo{number}{5} (\bibinfo{year}{2021}), \bibinfo{pages}{2845--2855}.
\newblock


\bibitem[Pan et~al\mbox{.}(2023a)]%
        {GaitRec}
\bibfield{author}{\bibinfo{person}{Honghu Pan}, \bibinfo{person}{Yongyong Chen}, \bibinfo{person}{Tingyang Xu}, \bibinfo{person}{Yunqi He}, {and} \bibinfo{person}{Zhenyu He}.} \bibinfo{year}{2023}\natexlab{a}.
\newblock \showarticletitle{Toward complete-view and high-level pose-based gait recognition}.
\newblock \bibinfo{journal}{\emph{IEEE Transactions on Information Forensics and Security}}  \bibinfo{volume}{18} (\bibinfo{year}{2023}), \bibinfo{pages}{2104--2118}.
\newblock


\bibitem[Pan et~al\mbox{.}(2023b)]%
        {ReID}
\bibfield{author}{\bibinfo{person}{Honghu Pan}, \bibinfo{person}{Qiao Liu}, \bibinfo{person}{Yongyong Chen}, \bibinfo{person}{Yunqi He}, \bibinfo{person}{Yuan Zheng}, \bibinfo{person}{Feng Zheng}, {and} \bibinfo{person}{Zhenyu He}.} \bibinfo{year}{2023}\natexlab{b}.
\newblock \showarticletitle{Pose-aided video-based person re-identification via recurrent graph convolutional network}.
\newblock \bibinfo{journal}{\emph{IEEE Transactions on Circuits and Systems for Video Technology}} \bibinfo{volume}{33}, \bibinfo{number}{12} (\bibinfo{year}{2023}), \bibinfo{pages}{7183--7196}.
\newblock


\bibitem[Pan et~al\mbox{.}(2024)]%
        {UCIG}
\bibfield{author}{\bibinfo{person}{Honghu Pan}, \bibinfo{person}{Wenjie Pei}, \bibinfo{person}{Xin Li}, {and} \bibinfo{person}{Zhenyu He}.} \bibinfo{year}{2024}\natexlab{}.
\newblock \showarticletitle{Unified Conditional Image Generation for Visible-Infrared Person Re-Identification}.
\newblock \bibinfo{journal}{\emph{IEEE Transactions on Information Forensics and Security}} (\bibinfo{year}{2024}).
\newblock


\bibitem[Parmar et~al\mbox{.}(2022)]%
        {CleanFID}
\bibfield{author}{\bibinfo{person}{Gaurav Parmar}, \bibinfo{person}{Richard Zhang}, {and} \bibinfo{person}{Jun-Yan Zhu}.} \bibinfo{year}{2022}\natexlab{}.
\newblock \showarticletitle{On Aliased Resizing and Surprising Subtleties in GAN Evaluation}. In \bibinfo{booktitle}{\emph{CVPR}}.
\newblock


\bibitem[Ren et~al\mbox{.}(2015)]%
        {FasterRCNN}
\bibfield{author}{\bibinfo{person}{Shaoqing Ren}, \bibinfo{person}{Kaiming He}, \bibinfo{person}{Ross Girshick}, {and} \bibinfo{person}{Jian Sun}.} \bibinfo{year}{2015}\natexlab{}.
\newblock \showarticletitle{Faster r-cnn: Towards real-time object detection with region proposal networks}.
\newblock \bibinfo{journal}{\emph{Advances in neural information processing systems}}  \bibinfo{volume}{28} (\bibinfo{year}{2015}).
\newblock


\bibitem[Rombach et~al\mbox{.}(2021)]%
        {StableDiffusion}
\bibfield{author}{\bibinfo{person}{Robin Rombach}, \bibinfo{person}{Andreas Blattmann}, \bibinfo{person}{Dominik Lorenz}, \bibinfo{person}{Patrick Esser}, {and} \bibinfo{person}{Björn Ommer}.} \bibinfo{year}{2021}\natexlab{}.
\newblock \bibinfo{title}{High-Resolution Image Synthesis with Latent Diffusion Models}.
\newblock
\newblock
\showeprint[arxiv]{2112.10752}~[cs.CV]


\bibitem[Ronneberger et~al\mbox{.}(2015)]%
        {UNet}
\bibfield{author}{\bibinfo{person}{Olaf Ronneberger}, \bibinfo{person}{Philipp Fischer}, {and} \bibinfo{person}{Thomas Brox}.} \bibinfo{year}{2015}\natexlab{}.
\newblock \showarticletitle{U-net: Convolutional networks for biomedical image segmentation}. In \bibinfo{booktitle}{\emph{Medical Image Computing and Computer-Assisted Intervention--MICCAI 2015: 18th International Conference, Munich, Germany, October 5-9, 2015, Proceedings, Part III 18}}. Springer, \bibinfo{pages}{234--241}.
\newblock


\bibitem[Saharia et~al\mbox{.}(2022)]%
        {Palette}
\bibfield{author}{\bibinfo{person}{Chitwan Saharia}, \bibinfo{person}{William Chan}, \bibinfo{person}{Huiwen Chang}, \bibinfo{person}{Chris Lee}, \bibinfo{person}{Jonathan Ho}, \bibinfo{person}{Tim Salimans}, \bibinfo{person}{David Fleet}, {and} \bibinfo{person}{Mohammad Norouzi}.} \bibinfo{year}{2022}\natexlab{}.
\newblock \showarticletitle{Palette: Image-to-image diffusion models}. In \bibinfo{booktitle}{\emph{ACM SIGGRAPH 2022 Conference Proceedings}}. \bibinfo{pages}{1--10}.
\newblock


\bibitem[Seitzer(2020)]%
        {PytorhcFID}
\bibfield{author}{\bibinfo{person}{Maximilian Seitzer}.} \bibinfo{year}{2020}\natexlab{}.
\newblock \bibinfo{title}{{pytorch-fid: FID Score for PyTorch}}.
\newblock \bibinfo{howpublished}{\url{https://github.com/mseitzer/pytorch-fid}}.
\newblock
\newblock
\shownote{Version 0.3.0}.


\bibitem[Song et~al\mbox{.}(2022)]%
        {DDIM}
\bibfield{author}{\bibinfo{person}{Jiaming Song}, \bibinfo{person}{Chenlin Meng}, {and} \bibinfo{person}{Stefano Ermon}.} \bibinfo{year}{2022}\natexlab{}.
\newblock \bibinfo{title}{Denoising Diffusion Implicit Models}.
\newblock
\newblock
\showeprint[arxiv]{2010.02502}~[cs.LG]


\bibitem[Theiss et~al\mbox{.}(2022)]%
        {VSAIT}
\bibfield{author}{\bibinfo{person}{Justin Theiss}, \bibinfo{person}{Jay Leverett}, \bibinfo{person}{Daeil Kim}, {and} \bibinfo{person}{Aayush Prakash}.} \bibinfo{year}{2022}\natexlab{}.
\newblock \showarticletitle{Unpaired image translation via vector symbolic architectures}. In \bibinfo{booktitle}{\emph{European Conference on Computer Vision}}. Springer, \bibinfo{pages}{17--32}.
\newblock


\bibitem[Vaswani et~al\mbox{.}(2017)]%
        {Transformer}
\bibfield{author}{\bibinfo{person}{Ashish Vaswani}, \bibinfo{person}{Noam Shazeer}, \bibinfo{person}{Niki Parmar}, \bibinfo{person}{Jakob Uszkoreit}, \bibinfo{person}{Llion Jones}, \bibinfo{person}{Aidan~N Gomez}, \bibinfo{person}{{\L}ukasz Kaiser}, {and} \bibinfo{person}{Illia Polosukhin}.} \bibinfo{year}{2017}\natexlab{}.
\newblock \showarticletitle{Attention is all you need}.
\newblock \bibinfo{journal}{\emph{Advances in neural information processing systems}}  \bibinfo{volume}{30} (\bibinfo{year}{2017}).
\newblock


\bibitem[Wang et~al\mbox{.}(2022)]%
        {DiffusionGAN}
\bibfield{author}{\bibinfo{person}{Zhendong Wang}, \bibinfo{person}{Huangjie Zheng}, \bibinfo{person}{Pengcheng He}, \bibinfo{person}{Weizhu Chen}, {and} \bibinfo{person}{Mingyuan Zhou}.} \bibinfo{year}{2022}\natexlab{}.
\newblock \showarticletitle{Diffusion-GAN: Training GANs with Diffusion}. In \bibinfo{booktitle}{\emph{The Eleventh International Conference on Learning Representations}}.
\newblock


\bibitem[Zhang et~al\mbox{.}(2022)]%
        {DINO}
\bibfield{author}{\bibinfo{person}{Hao Zhang}, \bibinfo{person}{Feng Li}, \bibinfo{person}{Shilong Liu}, \bibinfo{person}{Lei Zhang}, \bibinfo{person}{Hang Su}, \bibinfo{person}{Jun Zhu}, \bibinfo{person}{Lionel~M. Ni}, {and} \bibinfo{person}{Heung-Yeung Shum}.} \bibinfo{year}{2022}\natexlab{}.
\newblock \bibinfo{title}{DINO: DETR with Improved DeNoising Anchor Boxes for End-to-End Object Detection}.
\newblock
\newblock
\showeprint[arxiv]{2203.03605}~[cs.CV]


\bibitem[Zhang et~al\mbox{.}(2021)]%
        {VFNet}
\bibfield{author}{\bibinfo{person}{Haoyang Zhang}, \bibinfo{person}{Ying Wang}, \bibinfo{person}{Feras Dayoub}, {and} \bibinfo{person}{Niko Sunderhauf}.} \bibinfo{year}{2021}\natexlab{}.
\newblock \showarticletitle{Varifocalnet: An iou-aware dense object detector}. In \bibinfo{booktitle}{\emph{Proceedings of the IEEE/CVF conference on computer vision and pattern recognition}}. \bibinfo{pages}{8514--8523}.
\newblock


\bibitem[Zhang et~al\mbox{.}(2018)]%
        {LPIPS}
\bibfield{author}{\bibinfo{person}{Richard Zhang}, \bibinfo{person}{Phillip Isola}, \bibinfo{person}{Alexei~A Efros}, \bibinfo{person}{Eli Shechtman}, {and} \bibinfo{person}{Oliver Wang}.} \bibinfo{year}{2018}\natexlab{}.
\newblock \showarticletitle{The Unreasonable Effectiveness of Deep Features as a Perceptual Metric}. In \bibinfo{booktitle}{\emph{CVPR}}.
\newblock


\bibitem[Zhao et~al\mbox{.}(2022)]%
        {EGSDE}
\bibfield{author}{\bibinfo{person}{Min Zhao}, \bibinfo{person}{Fan Bao}, \bibinfo{person}{Chongxuan Li}, {and} \bibinfo{person}{Jun Zhu}.} \bibinfo{year}{2022}\natexlab{}.
\newblock \showarticletitle{Egsde: Unpaired image-to-image translation via energy-guided stochastic differential equations}.
\newblock \bibinfo{journal}{\emph{Advances in Neural Information Processing Systems}}  \bibinfo{volume}{35} (\bibinfo{year}{2022}), \bibinfo{pages}{3609--3623}.
\newblock


\bibitem[Zheng et~al\mbox{.}(2015)]%
        {Market151}
\bibfield{author}{\bibinfo{person}{Liang Zheng}, \bibinfo{person}{Liyue Shen}, \bibinfo{person}{Lu Tian}, \bibinfo{person}{Shengjin Wang}, \bibinfo{person}{Jingdong Wang}, {and} \bibinfo{person}{Qi Tian}.} \bibinfo{year}{2015}\natexlab{}.
\newblock \showarticletitle{Scalable Person Re-identification: A Benchmark}. In \bibinfo{booktitle}{\emph{Computer Vision, IEEE International Conference on}}.
\newblock


\bibitem[Zheng et~al\mbox{.}(2017)]%
        {PRW}
\bibfield{author}{\bibinfo{person}{Liang Zheng}, \bibinfo{person}{Hengheng Zhang}, \bibinfo{person}{Shaoyan Sun}, \bibinfo{person}{Manmohan Chandraker}, \bibinfo{person}{Yi Yang}, {and} \bibinfo{person}{Qi Tian}.} \bibinfo{year}{2017}\natexlab{}.
\newblock \showarticletitle{Person re-identification in the wild}. In \bibinfo{booktitle}{\emph{Proceedings of the IEEE conference on computer vision and pattern recognition}}. \bibinfo{pages}{1367--1376}.
\newblock


\bibitem[Zheng et~al\mbox{.}(2023)]%
        {TISemantic}
\bibfield{author}{\bibinfo{person}{Yu Zheng}, \bibinfo{person}{Fugen Zhou}, \bibinfo{person}{Shangying Liang}, \bibinfo{person}{Wentao Song}, {and} \bibinfo{person}{Xiangzhi Bai}.} \bibinfo{year}{2023}\natexlab{}.
\newblock \showarticletitle{Semantic segmentation in thermal videos: a new benchmark and multi-granularity contrastive learning-based framework}.
\newblock \bibinfo{journal}{\emph{IEEE Transactions on Intelligent Transportation Systems}} (\bibinfo{year}{2023}).
\newblock


\bibitem[Zhou et~al\mbox{.}(2023)]%
        {DDBM}
\bibfield{author}{\bibinfo{person}{Linqi Zhou}, \bibinfo{person}{Aaron Lou}, \bibinfo{person}{Samar Khanna}, {and} \bibinfo{person}{Stefano Ermon}.} \bibinfo{year}{2023}\natexlab{}.
\newblock \showarticletitle{Denoising diffusion bridge models}.
\newblock \bibinfo{journal}{\emph{arXiv preprint arXiv:2309.16948}} (\bibinfo{year}{2023}).
\newblock


\bibitem[Zhu et~al\mbox{.}(2017)]%
        {CycleGAN}
\bibfield{author}{\bibinfo{person}{Jun-Yan Zhu}, \bibinfo{person}{Taesung Park}, \bibinfo{person}{Phillip Isola}, {and} \bibinfo{person}{Alexei~A Efros}.} \bibinfo{year}{2017}\natexlab{}.
\newblock \showarticletitle{Unpaired image-to-image translation using cycle-consistent adversarial networks}. In \bibinfo{booktitle}{\emph{Proceedings of the IEEE international conference on computer vision}}. \bibinfo{pages}{2223--2232}.
\newblock


\end{thebibliography}

\end{document}


\title{Supplementary Materials}







\maketitle

\section{The sampling parameters of DPM-solver++}
Table~\ref{Tab:DPMparams} shows the hyperparameters of DPM-solver++~\cite{DPMSolverPP} used for speeding up sampling our Edge-guided Conditional Diffusion Model (ECDM) in the Two-stage Modality Adversarial Training (TMAT) strategy. We utilize DPM-solver++ solely during the training phase.

\begin{table}[htb]
  \centering
  \caption{The parameters in dpm-solver++}\label{Tab:DPMparams}
  \begin{tabular}{ll}
  \toprule
  Hyper-parameters          & Value        \\ 
  \midrule
  timesteps           & 5            \\
  order           & 3            \\
  skip type       & time uniform \\
  sampling method          & adaptive     \\
  type            & taylor       \\
  condition scale & 0.5          \\
  absolute tolerance            & 0.0078       \\
  relative tolerance           & 0.05         \\ 
  \bottomrule
  \end{tabular}
\end{table}

\section{Exploration of thermal image generation using different conditions on diffusion model}
We conducte an evaluation of the quality of generated images under various conditions. Our report includes metrics for the quality of generated images under the following conditions: no condition (marked as NO), thermal image condition (marked as $\mathcal{D} _{llvip}^{tir}$), thermal edge condition (marked as $\zeta_{llvip}^{tir}$), visible image in nighttime (marked as $\mathcal{D} _{llvip}^{vis}$), visible edge image in nighttime (marked as $\zeta_{llvip}^{vis}$), visible image in daytime (marked as $\mathcal{D} _{prw}$), and visible edge images in daytime (marked as $\zeta_{prw}$). $\mathcal{D} _{llvip}^{vis}$ and $\zeta_{llvip}^{tir}$ have the same distribution with the thermal domain. $\mathcal{D} _{llvip}^{vis}$ and $\zeta_{llvip}^{vis}$ is strictly aligned in time and space with $\mathcal{D}_{llvip}^{vis}$ and $\zeta_{llvip}^{tir}$, so it has different distribution with thermal domain but has same semantic information. $\mathcal{D} _{prw}$ and $\zeta_{prw}$ neither has same distribution nor semantic information. 
\begin{figure}[t]
  \centering
  \includegraphics[width=3.3in]{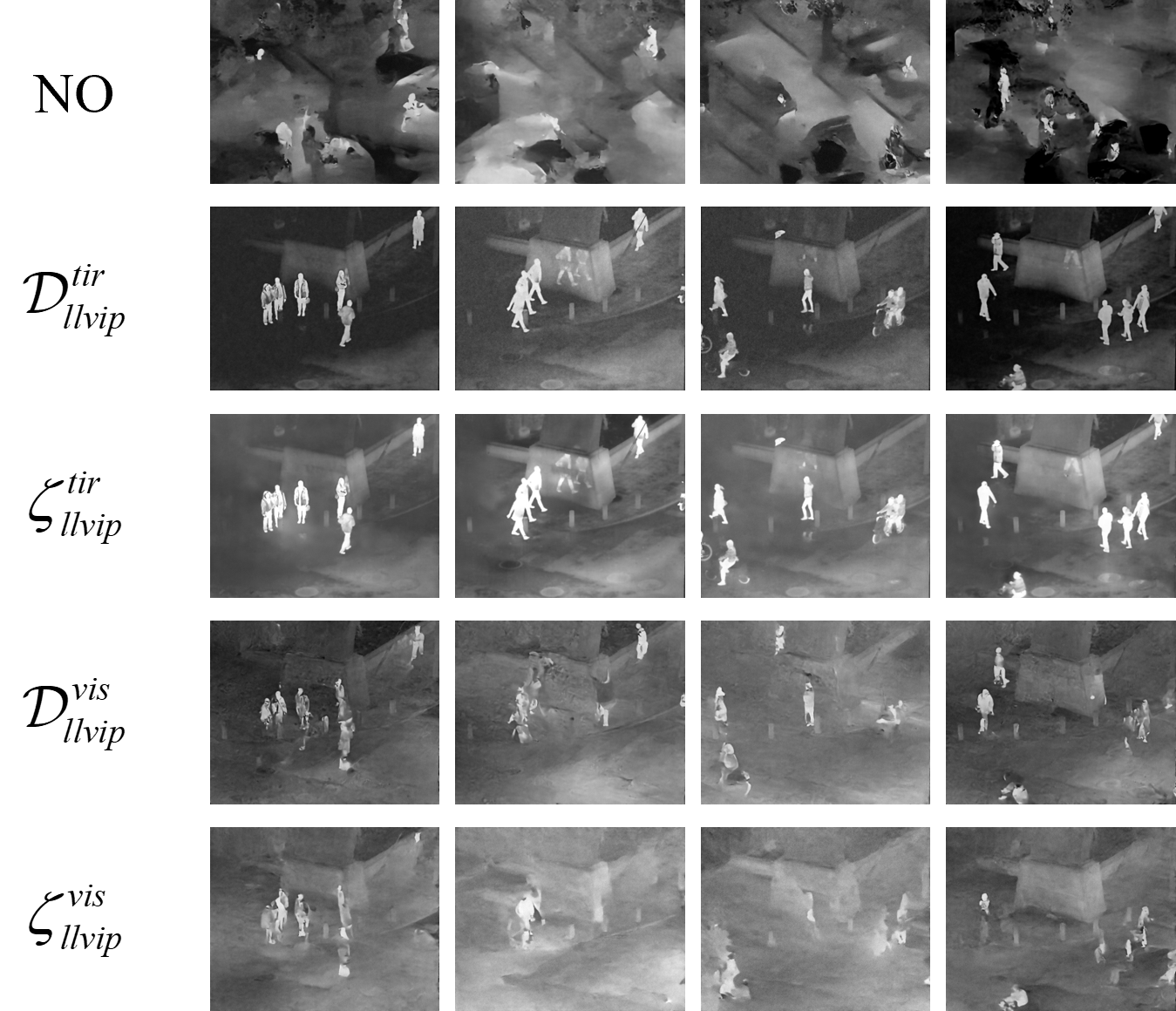}
  \caption{Visualization of images generated under different conditions.}\label{Fig:visual}
\end{figure}
\begin{table}[!htb]
    \centering
    \caption{Ablation study for different conditions}\label{Tab:condition}
    \resizebox{3.3in}{!}{
      \begin{tabular}{l cccc}
      \toprule
      Condition & FID$\downarrow$ & FID-C$\downarrow$ & FID-C$_{clip}\downarrow$ & KID$\downarrow$ \\
      \midrule
      NO&  257.14&250.82&38.59&0.2817\\
      $\mathcal{D} _{llvip}^{tir}$&  67.64&62.53&15.17&0.0408\\
      $\zeta_{llvip}^{tir}$&  35.07&35.69&16.15&0.0193\\
      $\mathcal{D} _{llvip}^{vis}$&  130.91&133.53&26.53&0.0967\\
      $\zeta_{llvip}^{vis}$&  139.91&147.09&26.98&0.1167\\
      \bottomrule
    \end{tabular}
    }
\end{table}

In this experiment, we maintain the training setting identical to the sampling condition. To ensure a fair comparison, we set $S_{diff}=70$. The results are presented in Table~\ref{Tab:condition} and visualized in Figure~\ref{Fig:visual}. When no condition is applied to control the generated content, the generated images exhibit a high FID-C score of 250.82 and lack meaningful content. By incorporating conditions, we observe a significant reduction in the FID-C score and improved control over the generated image content. Thermal domain conditions outperform visible domain conditions due to their similar distribution with the target domain. Notably, $\zeta_{llvip}^{tir}$ performs better than $\mathcal{D} _{llvip}^{vis}$, as the texture information in $\mathcal{D} {llvip}^{vis}$ adversely affects the fine control of edge information in the generated image boundaries. However, $\zeta{llvip}^{vis}$ performs relatively poorly compared to $\mathcal{D} _{llvip}^{vis}$, since the visible images in the LLVIP dataset~\cite{LLVIP} are captured at night, resulting in scarce edge information in these images. This finding verifies the importance of edge information in precisely generating fine-granularity content of objects.

\section{More showcases of ECDM on thermal object detection}
\begin{figure}[ht]
    \centering
    \includegraphics[width=3.3in]{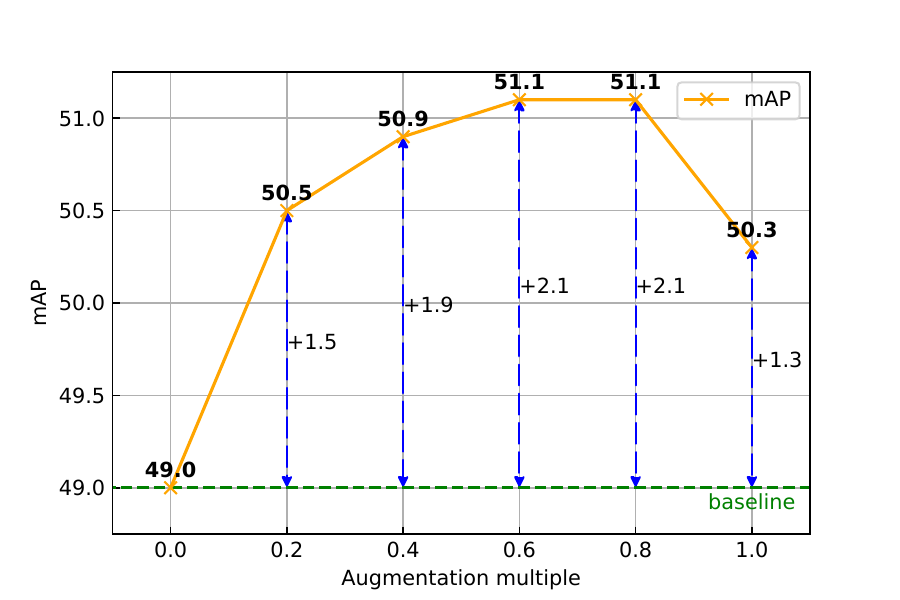}
    \caption{The performance of Faster RCNN trained with various amounts of generated pseudo training data. The x-axis indicates the augmentation multiple. For example, 0.2 indicates that the generated pseudo training data in the entire training sample is only 20\% of the real data.}\label{Fig:augFaster}
\end{figure}

\begin{figure}[!h]
    \centering
    \includegraphics[width=3.3in]{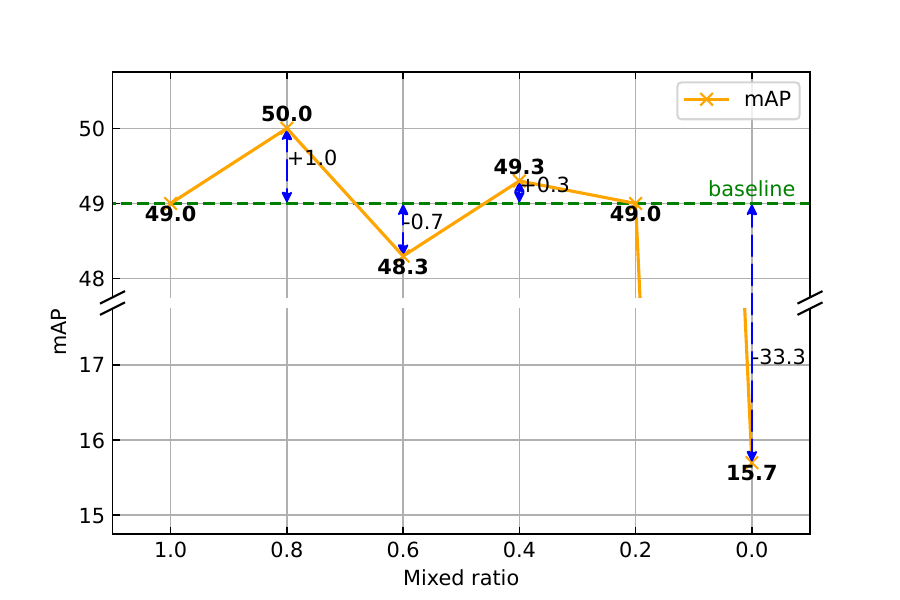}
    \caption{The performance of Faster RCNN trained with various amounts of generated pseudo training data. The x-axis indicates the mixed ratios. For example, 0.2 indicates that the entire training samples have 20\% generated pseudo training data and 80\% real data. }\label{Fig:mixedFaster}
\end{figure}

We also train Faster RCNN~\cite{FasterRCNN} with diverse augmentation multiple ratios and mixed ratios. The results are shown in Figures~\ref{Fig:augFaster} and~\ref{Fig:mixedFaster}, respectively. Note that the best augmentation multiple ratio at 0.8 in Faster RCNN is 0.8, which achieves a 2.1 improvement on mAP.

\begin{table*}[h]
    \centering
  \caption{Class-wise mAP results on the FLIR dataset}\label{Tab:submAP}
    \resizebox{6.6in}{!}{
  \begin{tabular}{l l c llllll}
  \toprule
  Method &Class &Pseudo data & mAP &AP@50    &AP@75&AP$_{s}$&     AP$_{m}$ & AP$_{l}$       \\ 
  \midrule
  \multirow{10}{*}{Faster RCNN}& \multirow{2}{*}{Person}&\ding{56}&26.1&50.4&24.5&19.1&53.9&54.3 \\
  &&\ding{52}&26.8 \textcolor{red}{(+0.7)}&51.7 \textcolor{red}{(+1.3)}&25.1 \textcolor{red}{(+0.6)}&20.1 \textcolor{red}{(+1.0)}&54.5 \textcolor{red}{(+0.6)}&57.5 \textcolor{red}{(+3.2)} \\
  \cmidrule{2-9}

  & \multirow{2}{*}{Bike}&\ding{56}&22.3&43.6&20.2&12.1&31.7&35.3 \\
  &&\ding{52}&25.7 \textcolor{red}{(+3.4)}&45.1 \textcolor{red}{(+1.5)}&28.0 \textcolor{red}{(+7.8)}&11.8 \textcolor{green}{(-0.3)}&37.8 \textcolor{red}{(+6.1)}&25.2 \textcolor{green}{(-0.1)} \\
  \cmidrule{2-9}

  & \multirow{2}{*}{Car}&\ding{56}&43.1&66.9&46.6&23.7&66.8&83.6 \\
  &&\ding{52}&46.0 \textcolor{red}{(+2.9)}&69.9&50.1 \textcolor{red}{(+3.5)}&26.6 \textcolor{red}{(+2.9)}&68.9 \textcolor{red}{(+2.1)}&84.5 \textcolor{red}{(+0.9)} \\
  \cmidrule{2-9}

  & \multirow{2}{*}{Light}&\ding{56}&10.0&25.6&4.9&9.4&31.4&- \\
  &&\ding{52}&10.7 \textcolor{red}{(+0.7)}&26.8 \textcolor{red}{(+1.2)}&6.5 \textcolor{red}{(+1.6)}&10.0 \textcolor{red}{(+0.6)}&33.9 \textcolor{red}{(+2.5)}&- \\
  \cmidrule{2-9}

  & \multirow{2}{*}{Sign}&\ding{56}&14.6&24.7&16.3&12.4&42.8&- \\
  &&\ding{52}&16.6 \textcolor{red}{(+2.0)}&27.8 \textcolor{red}{(+3.1)}&17.9 \textcolor{red}{(+1.6)}&13.8 \textcolor{red}{(+1.4)}&50.3 \textcolor{red}{(+7.5)}&- \\
  \midrule

  \multirow{10}{*}{RetinaNet}& \multirow{2}{*}{Person}&\ding{56}&14.8&37.8&8.9&7.6&40.5&46.8 \\
  &&\ding{52}&16.1 \textcolor{red}{(+1.3)}&39.6 \textcolor{red}{(+1.8)}&10.9 \textcolor{red}{(+2.0)}&8.2 \textcolor{red}{(+0.6)}&44.0 \textcolor{red}{(+3.5)}&52.6 \textcolor{red}{(+5.8)} \\
  \cmidrule{2-9}

  & \multirow{2}{*}{Bike}&\ding{56}&14.5&33.7&10.1&5.5&23.0&40.4 \\
  &&\ding{52}&15.9 \textcolor{red}{(+1.4)}&36.1 \textcolor{red}{(+2.4)}&11.7 \textcolor{red}{(+1.6)}&4.3 \textcolor{green}{(-1.2)}&26.2 \textcolor{red}{(+3.2)}&45.4 \textcolor{red}{(+5.0)} \\
  \cmidrule{2-9}

  & \multirow{2}{*}{Car}&\ding{56}&35.5&57.9&36.6&9.9&64.4&81.7 \\
  &&\ding{52}&35.5&58.8 \textcolor{red}{(+0.9)}&36.7 \textcolor{red}{(+0.1)}&10.2 \textcolor{red}{(+0.3)}&64.4&81.3 \textcolor{green}{(-0.4)} \\
  \cmidrule{2-9}

  & \multirow{2}{*}{Light}&\ding{56}&2.6&7.2&1.3&1.5&24.8&- \\
  &&\ding{52}&2.3 \textcolor{green}{(-0.3)}&6.9 \textcolor{green}{(-0.3)}&1.2 \textcolor{green}{(-0.1)}&1.6 \textcolor{red}{(+0.1)}&24.6 \textcolor{green}{(-0.2)}&- \\
  \cmidrule{2-9}

  & \multirow{2}{*}{Sign}&\ding{56}&5.1&10.4&4.9&2.2&40.3& \\
  &&\ding{52}&5.9 \textcolor{red}{(+0.8)}&12.6 \textcolor{red}{(+2.2)}&5.4 \textcolor{red}{(+0.5)}&2.9 \textcolor{red}{(+0.7)}&42.4 \textcolor{red}{(+2.1)}&- \\
  \midrule

  \multirow{10}{*}{CenterNet}& \multirow{2}{*}{Person}&\ding{56}&26.2&57.5&20.8&20.4&53.1&53.7 \\
  &&\ding{52}&28.9 \textcolor{red}{(+2.7)}&60.3 \textcolor{red}{(+2.8)}&24.1 \textcolor{red}{(+3.3)}&22.2 \textcolor{red}{(+1.8)}&54.7 \textcolor{red}{(+1.6)}&59.5 \textcolor{red}{(+5.8)} \\
  \cmidrule{2-9}

  & \multirow{2}{*}{Bike}&\ding{56}&22.6&39.9&22.7&7.3&36.2&32.3 \\
  &&\ding{52}&25.5 \textcolor{red}{(+2.9)}&45.0 \textcolor{red}{(+5.1)}&23.7 \textcolor{red}{(+1.0)}&11.1 \textcolor{red}{(+3.8)}&37.5 \textcolor{red}{(+1.3)}&35.3 \textcolor{red}{(+3.0)} \\
  \cmidrule{2-9}

  & \multirow{2}{*}{Car}&\ding{56}&45.5&72.7&46.5&23.7&69.6&85.5 \\
  &&\ding{52}&47.6 \textcolor{red}{(+2.1)}&74.6 \textcolor{red}{(+1.9)}&49.2 \textcolor{red}{(+2.7)}&26.3 \textcolor{red}{(+2.6)}&70.9 \textcolor{red}{(+1.3)}&85.8 \textcolor{red}{(+0.3)} \\
  \cmidrule{2-9}

  & \multirow{2}{*}{Light}&\ding{56}&14.7&39.1&6.6&14.1&38.1&- \\
  &&\ding{52}&15.7 \textcolor{red}{(+1.0)}&42.8 \textcolor{red}{(+2.7)}&7.7 \textcolor{red}{(+1.1)}&15.2 \textcolor{red}{(+1.1)}&37.2 \textcolor{green}{(-0.9)}&- \\
  \cmidrule{2-9}
  
  & \multirow{2}{*}{Sign}&\ding{56}&18.3&35.6&17.3&15.9&48.5&- \\
  &&\ding{52}&19.2 \textcolor{red}{(+0.9)}&38.0 \textcolor{red}{(+2.4)}&17.5 \textcolor{red}{(+0.2)}&16.6 \textcolor{red}{(+0.7)}&51.4 \textcolor{red}{(+2.9)}&- \\
  \midrule

  \multirow{10}{*}{VFNet}& \multirow{2}{*}{Person}&\ding{56}&16.0&40.2&10.2&10.3&40.0&40.6 \\
  &&\ding{52}&15.0 \textcolor{green}{(-1.0)}&37.9 \textcolor{green}{(-2.3)}&9.9 \textcolor{green}{(-0.3)}&8.7 \textcolor{green}{(-1.6)}&41.0 \textcolor{red}{(+1.0)}&44.4 \textcolor{red}{(+3.8)} \\
  \cmidrule{2-9}

  & \multirow{2}{*}{Bike}&\ding{56}&12.2&30.3&6.4&2.9&21.2&3.6 \\
  &&\ding{52}&10.2 \textcolor{green}{(-2.0)}&25.6 \textcolor{green}{(-4.7)}&5.7 \textcolor{green}{(-0.7)}&2.6 \textcolor{green}{(-0.3)}&17.6 \textcolor{green}{(-3.6)}&15.1 \textcolor{red}{(+11.5)} \\
  \cmidrule{2-9}

  & \multirow{2}{*}{Car}&\ding{56}&35.7&61.1&36.8&15.7&60.1&75.1 \\
  &&\ding{52}&32.5 \textcolor{green}{(-3.2)}&57.1 \textcolor{green}{(-4.0)}&33.3 \textcolor{green}{(-3.5)}&13.2 \textcolor{green}{(-2.5)}&56.1 \textcolor{green}{(-4.0)}&70.7 \textcolor{green}{(-4.4)} \\
  \cmidrule{2-9}

  & \multirow{2}{*}{Light}&\ding{56}&4.5&12.8&2.0&3.7&27.7&- \\
  &&\ding{52}&2.8 \textcolor{green}{(-1.7)}&8.0 \textcolor{green}{(-4.0)}&1.6 \textcolor{green}{(-0.4)}&2.3 \textcolor{green}{(-1.4)}&19.3 \textcolor{green}{(-8.4)}&- \\
  \cmidrule{2-9}

  & \multirow{2}{*}{Sign}&\ding{56}&7.2&15.8&5.6&4.9&36.4&- \\
  &&\ding{52}&4.4 \textcolor{green}{(-2.8)}&10.3 \textcolor{green}{(-5.5)}&3.1 \textcolor{green}{(-2.5)}&2.4 \textcolor{green}{(-2.5)}&29.7 \textcolor{green}{(-6.7)}&- \\
  \midrule

  \multirow{10}{*}{DINO}& \multirow{2}{*}{Person}&\ding{56}&12.4&28.2&9.0&11.0&18.5&21.2 \\
  &&\ding{52}&16.5 \textcolor{red}{(+4.1)}&39.3 \textcolor{red}{(+11.1)}&10.8 \textcolor{red}{(+1.8)}&13.4 \textcolor{red}{(+2.4)}&29.5 \textcolor{red}{(+11.0)}&32.2 \textcolor{red}{(+11.0)} \\
  \cmidrule{2-9}

  & \multirow{2}{*}{Bike}&\ding{56}&3.3&6.9&2.8&0.2&7.6&2.0 \\
  &&\ding{52}&3.5 \textcolor{red}{(+0.2)}&8.3 \textcolor{red}{(+1.4)}&1.6 \textcolor{green}{(-1.2)}&0.7 \textcolor{red}{(+0.5)}&6.3 \textcolor{red}{(+1.3)}&0.7 \textcolor{green}{(-1.3)} \\
  \cmidrule{2-9}

  & \multirow{2}{*}{Car}&\ding{56}&17.8&33.4&17.2&11.2&28.1&27.4 \\
  &&\ding{52}&20.2 \textcolor{red}{(+2.4)}&40.3 \textcolor{red}{(+6.9)}&18.0 \textcolor{red}{(+0.8)}&12.5 \textcolor{red}{(+1.3)}&32.2 \textcolor{red}{(+4.1)}&37.3 \textcolor{red}{(+9.9)} \\
  \cmidrule{2-9}

  & \multirow{2}{*}{Light}&\ding{56}&2.8&7.0&1.7&2.7&8.5&- \\
  &&\ding{52}&3.6 \textcolor{red}{(+0.8)}&9.2 \textcolor{red}{(+2.2)}&2.1 \textcolor{red}{(+0.4)}&3.5 \textcolor{red}{(+1.3)}&7.6 \textcolor{green}{(-0.9)}&- \\
  \cmidrule{2-9}

  & \multirow{2}{*}{Sign}&\ding{56}&2.2&4.6&1.7&2.0&7.0&- \\
  &&\ding{52}&2.5 \textcolor{red}{(+0.3)}&6.7 \textcolor{red}{(+2.1)}&1.3 \textcolor{green}{(-0.4)}&2.1 \textcolor{red}{(+0.1)}&8.2 \textcolor{red}{(+1.2)}&- \\
  \bottomrule
  \end{tabular}
  }
\end{table*}

\section{Class-wise results on the FLIR dataset}
We train various object detectors on the FLIR dataset~\cite{FLIR}, including Faster RCNN~\cite{FasterRCNN}, RetinaNet~\cite{RetinaNet}, CenterNet~\cite{CenterNet}, VFNet~\cite{VFNet}, and DINO~\cite{DINO}. For a fair comparison, we maintain an augmentation multiple ratio of 1.0 throughout this experiment. The FLIR dataset encompasses 15 categories, but we only utilize 5 categories in our experiments due to the limited labeling. The primary metrics of mAP are presented in the manuscript. We provide class-wise sub metrics of mAP in Table~\ref{Tab:submAP}.

\section{More qualitative results}
\begin{figure*}[htb]
    \centering
    \includegraphics[width=7in]{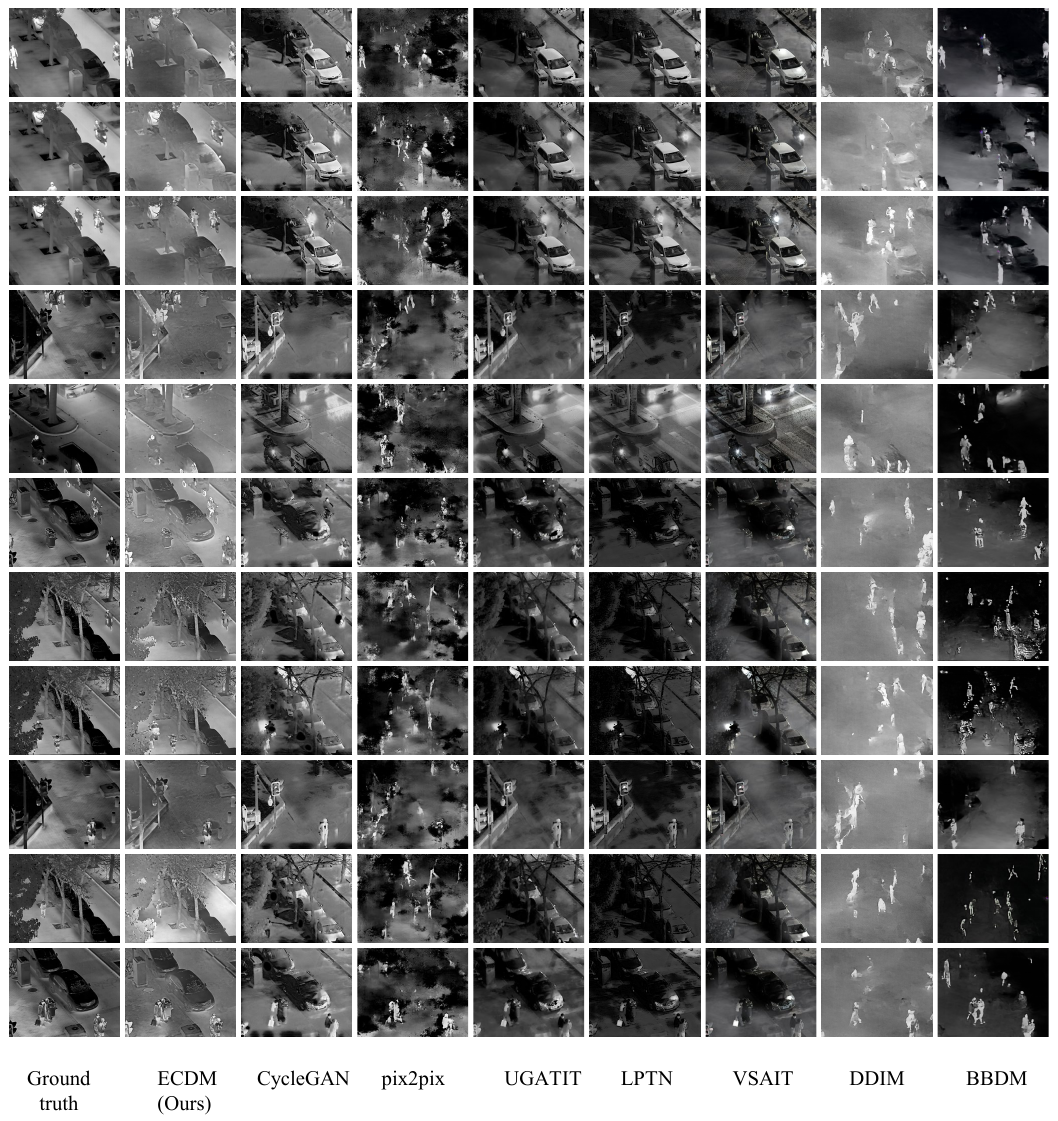}
    \caption{More qualitative comparison of our proposed method with other state-of-the-art methods on the LLVIP test dataset. To ensure fairness and randomness, we use Python's random module with a fixed seed (1234) to select images from the dataset. The selected images are `190065.jpg', `190072.jpg', `190127.jpg',`200143.jpg', `210307.jpg', `230422.jpg', `240321.jpg',
    `240409.jpg', `260211.jpg', `260304.jpg', `260379.jpg'.}\label{Fig:more_random_compare}
\end{figure*}

We provide more qualitative comparison results with other methods in Figure~\ref{Fig:more_random_compare}. 

\begin{figure}[htb]
    \centering
    \includegraphics[width=3.35in]{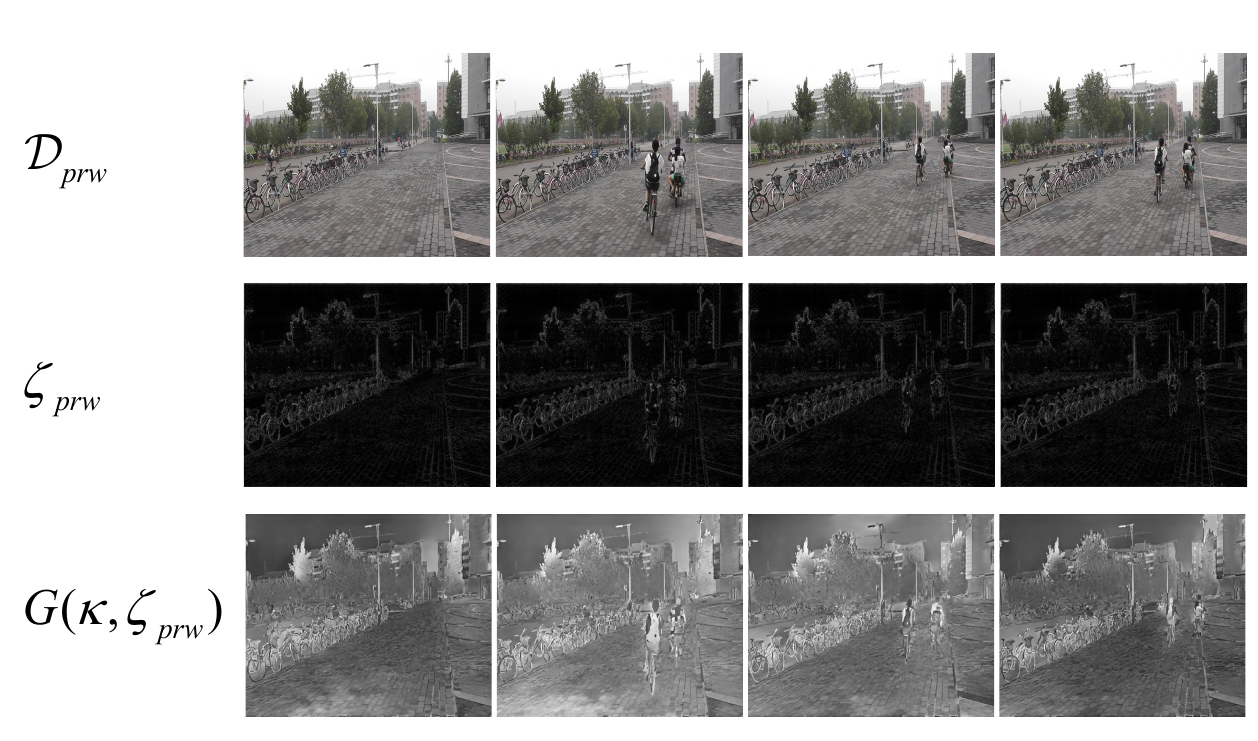}
    \caption{Here are some examples of images in the PRW dataset, edge images extracted from images and generated pseudo thermal images under edge images.}\label{Fig:genreratePRW}
\end{figure}

The generated samples under the PRW dataset are shown in Figure~\ref{Fig:genreratePRW}. Figure~\ref{Fig:genreratePRW} demonstrates that the generated thermal images exhibit similar overall gray distributions in the global space. However, some discrepancies are observed in specific details, such as the heads or legs of humans, and bags. These differences highlight the difficulty of the transferability models challenge, owing to the substantial gap in data distribution when generating infrared images from edge images sourced from different datasets.

\begin{figure}[ht]
    \centering
    \begin{subfigure}[b]{1.06in}
        \centering
        \includegraphics[width=1.06in]{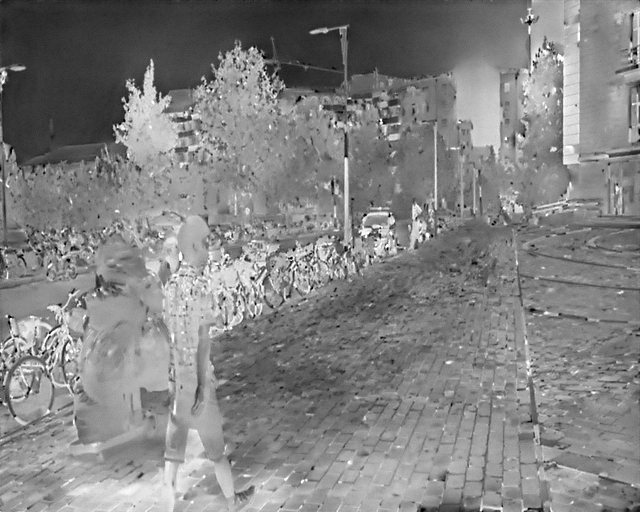}
        \caption{Blur ghost}\label{Fig:blur_ghost}
    \end{subfigure}
    \begin{subfigure}[b]{1.06in}
        \centering
        \includegraphics[width=1.06in]{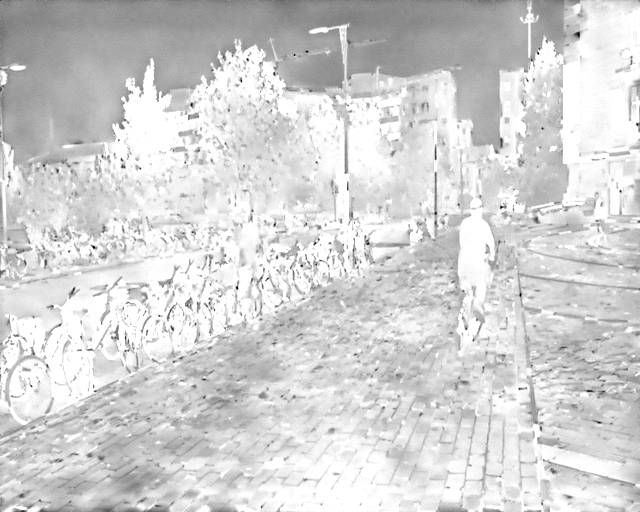}
        \caption{Error color levels}\label{Fig:error_color_level2}
    \end{subfigure}
    \begin{subfigure}[b]{1.06in}
        \centering
        \includegraphics[width=1.06in]{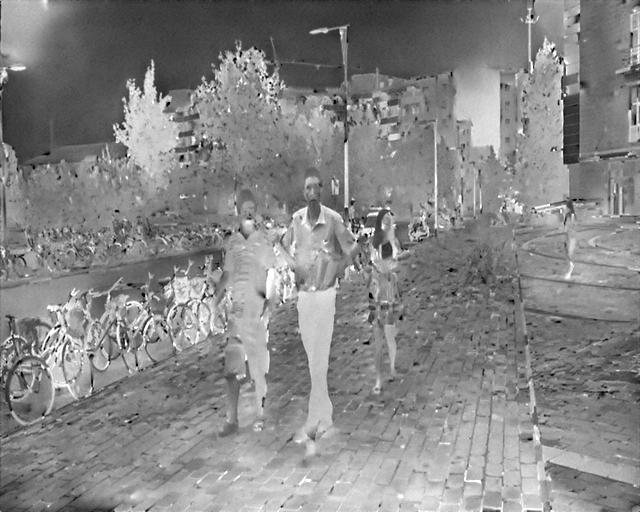}
        \caption{Polarity reversal}\label{Fig:polarity_reversal}
    \end{subfigure}
    \caption{Typical FAKE thermal images.\ (\subref{Fig:blur_ghost}) Blur ghost, which means exits some blurry artifacts in the images.\ (\subref{Fig:error_color_level2}) Error color levels, which means images have incorrect color levels.\ (\subref{Fig:polarity_reversal}) Polarity reversal, which means a hot object has a lower gray value than a cool object (face and cloth).
    }
    \label{Fig:fake}
\end{figure}

Some falied cases are shown in Figure~\ref{Fig:fake}.

\bibliographystyle{ACM-Reference-Format}
\bibliography{sample-base}